\documentclass[letterpaper]{article} 
\usepackage{aaai23}  
\usepackage{times}  
\usepackage{helvet}  
\usepackage{courier}  
\usepackage[hyphens]{url}  
\usepackage{graphicx} 
\usepackage{natbib}  
\usepackage{caption} 
\usepackage{algorithm}
\usepackage{algorithmic}

\usepackage{newfloat}
\usepackage{listings}
\DeclareCaptionStyle{ruled}{labelfont=normalfont,labelsep=colon,strut=off} 
\floatstyle{ruled}
\newfloat{listing}{tb}{lst}{}
\floatname{listing}{Listing}
\usepackage{placeins}
\usepackage{float}
\nocopyright

\usepackage{multirow}
\usepackage{bm}
\usepackage{mathtools}
\usepackage{caption}
\usepackage{subcaption}
\usepackage{lipsum}
\usepackage{booktabs}
\usepackage{float}

\newcommand{\Tau}{\mathcal{T}}
\DeclareMathOperator*{\argmax}{arg\,max}

\newcommand{\cut}{\textsc{Cut}}
\newcommand{\hole}{\textsc{Hole}}
\newcommand{\completionr}{\textsc{Completion-R}}
\newcommand{\completions}{\textsc{Completion-S}}
\newcommand{\uncompleted}{\textsc{Uncompleted}}

\newcommand{\pointnet}{\textsc{PointNet}}

\newcommand{\rgan}{r\textsc{-GAN}}

\newcommand{\treegan}{\textsc{TreeGAN}}
\newcommand{\betavae}{$\beta$\textsc{-VAE}}

\newcommand{\shapenet}{\textsc{ShapeNet}}

\newcommand{\spavae}{\textsc{SPA-VAE}}
\newcommand{\spavaeresampled}{\textsc{SPA-VAE-r}}

\newcommand{\editvae}{\textsc{EditVAE}}
\newcommand{\editvaeresampled}{\textsc{EditVAE-r}}

\newcommand{\balanced}{\textsc{Balanced}}

\title{\spavae : Similar-Parts-Assignment for \\
Unsupervised 3D Point Cloud Generation}
\author{
    Shidi Li\textsuperscript{\rm 1},
    Christian Walder\textsuperscript{\rm 1,2}\thanks{Now at Google Brain Montreal.},
    Miaomiao Liu\textsuperscript{\rm 1}
}
\affiliations{
    \textsuperscript{\rm 1}Australian National University\\
    \textsuperscript{\rm 2}Data61, CSIRO\\
    \{shidi.li, christian.walder, miaomiao.liu\}@anu.edu.au, 

}

\begin{document}

\maketitle

\begin{abstract}
    This paper addresses the problem of unsupervised parts-aware
    point cloud generation with learned parts-based self-similarity.
    Our \spavae\ infers a set of latent canonical candidate shapes for any given object, along with a set of rigid body transformations for each such candidate shape to one or more locations within the assembled object.
    In this way, noisy samples on the surface of, say, each leg of a table, are effectively combined to estimate a single leg prototype.  
    When parts-based self-similarity exists in the raw data, sharing data among parts in this way confers numerous advantages: modeling accuracy, appropriately self-similar generative outputs, precise in-filling of occlusions, and model parsimony.
    \spavae\ is trained end-to-end using a variational Bayesian approach which uses the Gumbel-softmax trick for shared part assignments, along with various novel losses to provide appropriate inductive biases.
    Quantitative and qualitative analyses on ShapeNet demonstrate the advantage of \spavae. 
\end{abstract}

\section{Introduction}

\nocite{pmlr-v80-walder18a}
\label{sec:intro}
Significant progress has been made recently in unsupervised 3D shape segmentation by using primitives\cite{tulsiani2017learning,paschalidou2019superquadrics,sun2019learning,yang2021unsupervised,paschalidou2020learning} and implicit representations\cite{chen2019bae,deng2020cvxnet,genova2019learning,genova2020local,paschalidou2021neural,li2021sp}, 
which has further boosted parts-based controllable generation~\cite{shu20193d} and editing~\cite{li2021editvae,li2021sp,yang2021cpcgan,gal2021mrgan}.
For example, recently~\citet{li2021editvae} achieved unsupervised parts-based controllable generation and editing by learning a disentangled (pose and shape) and dual (primitive and points) representation for 3D shapes.
However, existing parts-aware segmentation and generation approaches~\cite{paschalidou2020learning,li2021editvae} 
focus on parsing or modelling the object's parts in spatial locations
dependent on the category of the object. Due to the potentially imbalanced distribution of points across different parts of an object, existing methods suffer from inferior generation and segmentation results for parts with a small number of points, even when those parts are duplicated throughout the object, meaning that greater detail may be inferred.

\begin{figure}[ht]
    \centering
     \includegraphics[width=0.9\linewidth]{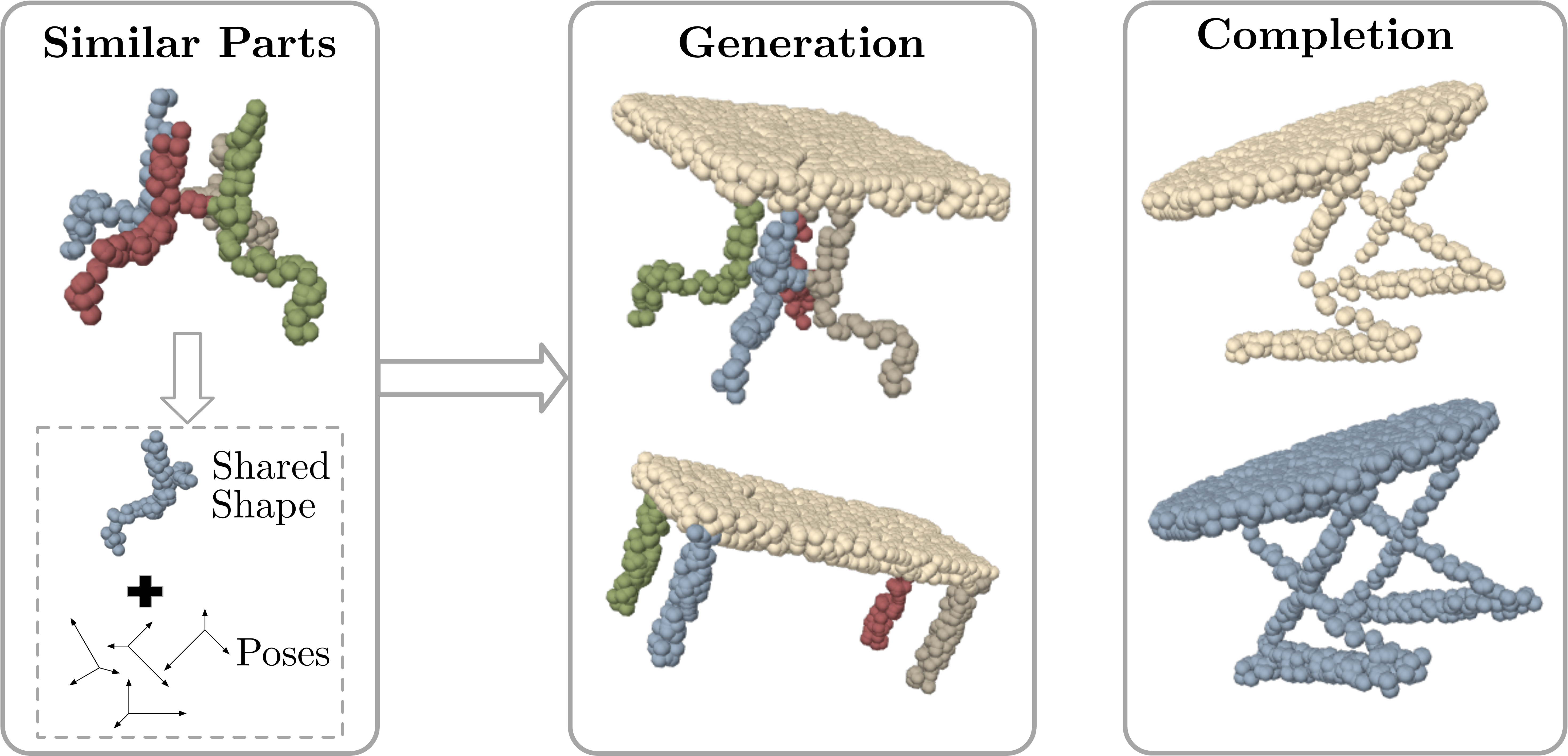}
  
     \caption{\spavae\ learns self-similarity by disentangling parts of the (unlabeled) 
     input point 
     cloud that have the same shape but different poses, allowing parts-aware generation and completion.
     With parts denoted by different colors,
     here we demonstrate: parts-aware \textit{generation} with similar table legs,
     and \textit{completion} (blue) via the inferred self-similarity of the incomplete input (yellow).}
     \label{fig:teaser}
\end{figure}
To address this problem, we propose to learn parts-based self-similarity in an
unsupervised manner, to infer a set (per object instance) of canonical shapes for point cloud
generation.
In particular, we model the similarity between parts by letting them share a~\emph{canonical
shape} instantiated with multiple poses. This is more general than a symmetry based notion of self-similarity 
wherein one part may be corresponded to another by mirroring across a particular hyperplane.
Rather, our notion of parts-based self-similarity needs only correspond (by rigid body transformation) one or more observed noisy or occluded parts to their object-instance-specific~\emph{canonical shape}.

In this paper, we propose \spavae , an unsupervised VAE-based framework for
point cloud generation which infers and exploits parts-based self-similarity. We seek to discover the~\emph{canonical shapes} underlying an object, to disentangle the observed parts' poses defined by rigid transformations relative to a canonical
shape coordinate system, and to model the associated joint distribution for generation. Our model disentangles poses, primitives, and points similarly to~\citet{li2021editvae}, by leveraging a specially designed disentanglement-oriented VAE framework~\cite{higgins2016beta,burgess2018understanding}. This yields part information at specific spatial locations. We further propose the 
Similar Parts Assignment (SPA) module as a mechanism for inferring a set of~\emph{canonical shapes}
for an object, along with an assignment matrix which associates each~\emph{canonical shape} with one or more poses, allowing to assemble the object.
To handle the non-differentiability of this hard assignment matrix, we apply the \textit{straight-through Gumbel-softmax} gradient approximation~\cite{jang2016categorical,maddison2016concrete}, which has been widely used in 3D computer vision~\cite{yang2019modeling}.
We further introduce a~\emph{diversity loss} to encourage our network to learn discriminative~\emph{canonical shapes}, and an~\emph{assignment loss} to guarantee the existence of~\emph{canonical shapes}. In addition to the quantitatively improved shape generation due to parts-based self-similarity learning, \spavae\  further yields intelligent shape completion at no extra cost.

In summary, our main contributions are as follows:
\begin{enumerate}
    \item We propose a framework for modelling parts-based self-similarity for
    3D shape modelling and generation.
    \item We provide a practical scheme based on the Gumbel-softmax trick for the non-differentiable parts assignment matrix, along with novel loss terms which we demonstrate to provide effective inductive biases.
    \item We demonstrate that our disentangled canonical shapes, assignment matrix and part poses improve point cloud generation and allow shape completion.
\end{enumerate}
Extensive qualitative and quantitative results on the popular \shapenet\ data set demonstrate the excellent performance that is obtainable by the novel similar-parts-assignment mechanism exemplified by our \spavae\ model.

\section{Related Work}
\label{sec:related_work}
\subsubsection{Parts-Aware 3D Point Cloud Generation.}
The topic of 3D point cloud generation has been widely studied in recent 
years~\cite{achlioptas2018learning,valsesia2018learning,hui2020progressive,wen2021learning,kim2020softflow,klokov2020discrete},
especially by learning deformation from vanilla point clouds or 
shapes~\cite{yang2019pointflow,cai2020learning,luo2021diffusion,zhou20213d,li2021sp}.
Another branch of work focuses on parts-aware point cloud generation with ground-truth 
semantic labels (indicating whether any given point belongs to \textit{e.g.} the base, back, 
or legs of a chair)~\cite{nash2017shape,mo2019structurenet,mo2020pt2pc,schor2019componet,yang2021cpcgan}.
As the requirement for well-aligned semantic part labels hinders real-world 
application, unsupervised parts-aware generation has been explored 
by~\citet{li2021editvae,shu20193d,li2021sp,postels2021go,gal2021mrgan}.
In particular, \citet{li2021sp} generates parts point clouds by utilizing the space 
consistent feature of the sphere, while \citet{shu20193d} organizes points by way of a 
latent tree structure,
and \citet{li2021editvae} further disentangled parts into their pose and shape, to achieve 
the controllable editing.
Finally, dozens of 3D shape reconstruction algorithms may potentially be applied to 
3D point cloud generation~\cite{chen2019bae,paschalidou2021neural,paschalidou2020learning,dubrovina2019composite}.
However, the above works ignore parts-based self-similarity, which may benefit 
fitting and generation by effectively sharing data for each part with the 
similar counterparts within the object (as in \textit{e.g.} the legs of a table).

\noindent\subsubsection{Unsupervised Self-Similarity Learning.}
Exploiting the self-symmetry or similarity property~\cite{huang2012point} has 
broad applications in image 
retrieval~\cite{yoon2020image,diao2021similarity,mishra2021effectively,xiao2021region,seo2021learning,wu2020unsupervised}, 
video-based action recognition~\cite{kwon2021learning},
music modelling~\cite{walder2018neural}
and depth map estimation~\cite{xu2020ladybird,zhou2021nerd}.
In particular, \citet{zhou2021nerd} detects planes of symmetry within input images in 
an unsupervised manner by building a 3D cost volume. While detecting the symmetric structure 
can improve the object geometry, such as the depth estimation, it is challenging to directly 
extend such approaches to
parts-aware object segmentation and generation due to the complexity of modeling 
mutual symmetries within groups of parts.
Our \spavae\ explores the self-similarity instead of self-symmetry for unsupervised learning 
of canonical shapes and object pose distributions, and contributes similar-parts-aware generation.

\subsubsection{3D Point Cloud Completion.}
The symmetry property has been explored in the 3D shape completion task in the 
literature~\cite{mitra2006partial,mitra2013symmetry,pauly2008discovering,podolak2006planar,sipiran2014approximate,sung2015data,thrun2005shape}.
Traditional geometry-based approaches require expensive optimizations and are sensitive 
to noise. Recent learning-based methods achieve 
superior performance especially in a supervised setting with paired complete 
shapes~\cite{xiang2021snowflakenet,xie2021style,gong2021me,wang2021voxel,wen2021pmp,huang2021rfnet,nie2020skeleton,xie2020grnet,wang2020cascaded,yuan2018pcn,tchapmi2019topnet}.
The lack of pairs of complete and incomplete data in real world settings, however, 
hinders the general applicability.
Recently, methods not based on such pairings have been proposed for the shape completion 
task~\cite{wang2020self,gu2020weakly,wen2021cycle4completion,chen2019unpaired,stutz2018learning,wu2020multimodal,zhang2021unsupervised}.
In particular, \citet{chen2019unpaired} maps reconstructed incomplete inputs to a latent 
space to imitate the complete one using GANs~\cite{goodfellow2014generative},
while the subsequent~\cite{zhang2021unsupervised} adopts GAN inversion~\cite{karras2019style,brock2018large} to find a best match from incomplete point clouds to reconstructed 
complete ones.
In comparison with existing methods, our model (which does not require pairs 
of incomplete and complete point clouds) merely leverages complete point 
clouds in training and incomplete ones in testing, by leveraging our inferred parts-based
self-similarities in an unsupervised manner.

\begin{figure*}[pt]
    \begin{center}
        \includegraphics[width=1\linewidth]{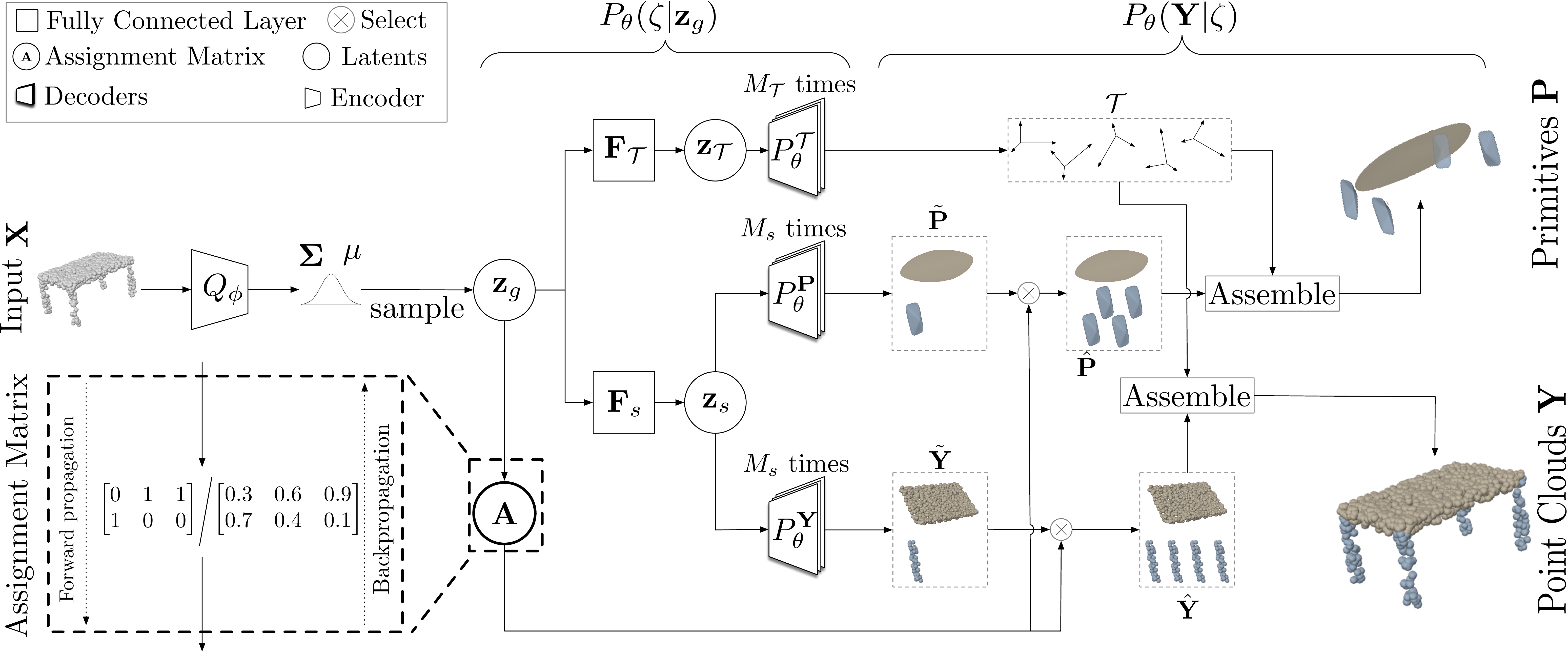}
    \end{center}
       \caption{An overview of the SPA-VAE architecture.
       Simplified from~\cite{li2021editvae}, the encoder $Q_\phi$ first infers the posterior from the input ground-truth point cloud $\bm X \in \bm R^{N\times 3}$.
       The global latent $\bm z_g$ is then sampled from the posterior using the reparameterization trick, 
       from which the shared shape latent $\bm z_s$ and pose latent $\bm z_\Tau$ are obtained from fully connected layers $F_s$ and $F_\Tau$, respectively.
       Further, poses and shapes (primitives and points) are reconstructed from the corresponding latents by pose decoders $P_\theta^{\bm \Tau}$, primitive decoders $P_\theta^{\bm P}$, and point decoders $P_\theta^{\bm Y}$, respectively.
       In another branch, the assignment matrix $\bm A$ is calculated as a function of the global latent $\bm z_g$, and represents parts similarity by denoting the sharing of shapes across multiple poses.
       Assigned poses and shapes are finally assembled to yield the reconstructed set of primitives $\bm P$ and point cloud $\bm Y$.
       }
    \label{fig:framework}
\end{figure*}

\section{Methodology}
\label{sec:methdology}
The focus of our SPA-VAE is on learning parts-based self-similarity for parts-aware 3D shape generation and completion --- see Fig.~\ref{fig:framework} for an architectural overview, and Appendix A of the supplementary material for an overview of our notation.
To learn the notion of self-similarity introduced in Sec.~\ref{sec:intro}, SPA-VAE disentangles 3D point clouds into the constituent pose and shape information of the parts therein, employing a disentanglement-oriented form of the Variational Auto-Encoder (VAE) framework~\cite{burgess2018understanding,higgins2016beta}.
Similar to~\citet{li2021editvae}, our resulting VAE-based model may be defined via an \textit{Evidence lower bound} (ELBO) objective~\cite{kingma2013auto,pmlr-v32-rezende14} as
\begin{align}\label{eq:vae}
    &\log P_\theta(X) \geq \\
    & \qquad \bm E_{ Q_\phi(\bm z_g \vert \bm X)}[\log P_\theta(\bm X \vert \bm \zeta)]
    - D_{KL}( Q_\phi(\bm z_g \vert \bm X) \Vert  P_\theta(\bm z_g))
    .
    \nonumber
\end{align}
The first r.h.s.\ term is the \textit{reconstruction error} and the second the \textit{variational regularizer}, while $ Q_\phi$ and $ P_\theta$ are the encoder and decoder, respectively.
Here $\bm\zeta$ denotes our \textit{disentangled} representation (defined shortly), and we approximate the posterior in $\bm\zeta$ by learning an approximate posterior $ Q_\phi(\bm z_g \vert \bm X)$ of a latent \textit{global} (non-disentangled) representation $\bm z_g$ via the ELBO. Following \cite{li2021editvae} we couple $\bm\zeta$ and $\bm z_g$ by way of a learnt deterministic mapping, \textit{i.e.} by choosing $ P_\theta(\bm\zeta \vert \bm z_g) =  Q_\phi(\bm\zeta \vert \bm z_g) \rightarrow \delta(\bm\zeta - \text{NN}_\theta(\bm z_g))$,
with $\delta(\cdot)$ as Dirac distribution and $\text{NN}_\theta$ be a neural network model.
As a result, we may define $\bm\zeta=\text{NN}_\theta(\bm z_g)$, which denotes the 3D shape corresponding to the global latent $\bm z_g$.
The 3D shape is further decomposed into $M_\Tau$ constituent parts as $\bm\zeta = \bigcup_{m=1}^{M_\Tau}\bm\zeta^m$. Each part is also decomposed further, as 
\begin{align}
\label{eqn:mthpart}
\bm\zeta^m = \{ \bm \Tau^m, \hat{\bm Y}^m, \hat{\bm P}^m \},    
\end{align}
which represents the pose $\bm \Tau^m$, primitive $\hat{\bm P}^m$, and points $\hat{\bm Y}^m$. We also --- particularly in Fig.~\ref{fig:framework} --- use the notation $\hat{\bm Y} = \bigcup_{m=1}^{M_\Tau}\hat{\bm Y}^m$, \textit{etc.}
The three components which make up a single part are defined as follows.

\subsubsection{Pose $\bm \Tau^m$} includes translation $\bm t \in \bm {R}^{3}$ and rotation 
$\bm R(\bm q) \in \bm {SO}(3)$.
Rotation is further defined by the quaternion $\bm q \in \bm R^{3}$; for details see~\cite{li2021editvae}.
These parameters represent the \textit{linear} transformation of a single part from canonical pose to object pose as $\bm x \mapsto \bm\Tau(\bm x)\equiv \bm R(\bm q)\bm x + \bm t$. 

\subsubsection{Primitive $\hat{\bm P}^m$} is deformed \textit{linearly} and \textit{non-linearly} from simple shapes.
We employ the superquadric parametrization with parameters $\eta$ and $\omega$ as in~\cite{paschalidou2019superquadrics,li2021editvae}.
Thus, the surface point is defined (in canonical coordinates) by:
\begin{equation}\label{eq:primitive}
    \begin{aligned}
    \bm r(\eta, \omega) = 
    \begin{bmatrix}
    \alpha_x \cos^{\epsilon_1}{\eta} \cos^{\epsilon_2}{\omega} \\
    \alpha_y \cos^{\epsilon_1}{\eta} \sin^{\epsilon_2}{\omega} \\
    \alpha_z \sin^{\epsilon_1}{\eta}
    \end{bmatrix}
    \begin{matrix}
    -\pi/2 \leq \eta \leq \pi/2 \\
    -\pi \leq \omega \leq \pi
    \end{matrix},
    \end{aligned}
\end{equation}
where $\bm \alpha = (\alpha_x, \alpha_y, \alpha_z)^\top$ and $\bm \epsilon = (\epsilon_1, \epsilon_2)^\top$ are the scale and shape parameters, respectively.
We further include two \textit{non-linear} deformation parameters as in~\cite{li2021editvae,barr1987global,paschalidou2019superquadrics}.

\subsubsection{Points $\hat{\bm Y}^m $} are 3D point clouds representing a single part in canonical coordinates with dimensionality $\bm R^{N_p\times 3}$, so that $N_p$ is number of points per part.

\subsection{Similar Parts Assignment (SPA)}\label{subsec:meth:assign_matrix}
To achieve self-similarity within the assembled object, we share each of a set of $M_s$ \textit{shapes} (each comprising a primitive and point cloud) across $M_\Tau\geq M_s$ \textit{parts} (each with unique pose). Shapes are not fixed across objects, but rather are latent variables which form part of the generation process for a given object. This is achieved by modelling
\begin{itemize}
    \item a set of $M_s$ latent \textit{shapes} defined by their primitives $\tilde{\bm P}=\bigcup_{m=1}^{M_s} \tilde{\bm P}^m$ along with the corresponding point clouds $\tilde{\bm Y}=\bigcup_{m=1}^{M_s} \tilde{\bm Y}^m$,
    \item an assignment matrix $\bm A \in \{0,1\}^{M_s\times M_\Tau}$ which associates each of the $M_\Tau$ latent part transformations $\bm \Tau^1, \dots, \bm\Tau^{M_\Tau}$ with exactly one of $M_s$ latent shapes.
\end{itemize}

The $j$-th column $\bm A_j$ of $\bm A$ is a one-hot vector which indicates which shape is coupled with the $j$-th pose. Precisely, this means that if $A_{ij}=1$, then 
\begin{align}
    \tilde{\bm P}^i = \hat{\bm P}^j \text{~~~and~~~}
    \tilde{\bm Y}^i = \hat{\bm Y}^j,
\end{align}
where we emphasize that \textit{e.g.} $\tilde{\bm P}^i$ represents the $i$-th latent shape primitive, while $\hat{\bm P}^j$ denotes the $j$-th primitive appearing in the object as per Eq.~\ref{eqn:mthpart} (and therefore coupled with the $j$-th pose $\bm \Tau^j$) --- see the notation guide in Appendix A.

To handle the categorical assignment matrix $\bm A$ we employ the \textit{Gumbel-softmax trick} and associated \textit{straight-through} gradient estimator~\cite{maddison2016concrete,jang2016categorical}. This means that $\bm A_j$ is defined as, for $j=1,2,\dots,M_\Tau$,
\begin{align}
\label{eq:assign_hard}
        \bm A_j = \mathrm{one\_hot}\big(\argmax_i \bm\Delta_\theta(\bm z_g)_j\big),
\end{align} 
where $\bm\Delta_\theta(\cdot)$ is a neural network module which maps the global latent $\bm z_g$ to a matrix of dimensionality $\bm R^{M_s\times M_\Tau}$,
while $\mathrm{one\_hot}(\cdot)$ maps the index returned by the $\argmax$ function to $\{0,1\}$ indicator form.

Because $\bm A$ is not differentiable w.r.t $\theta$, we employ the straight-through estimator on the soft assignment
\begin{align}\label{eq:assign_soft}
        \tilde{\bm A}_j = \bm\sigma\left(\frac{\log(\bm\Delta_\theta(\bm z_g)_j)+\bm g_j}{\tau}\right),
\end{align}
for some fixed $\tau>0$ where each $\bm g_j \in \bm R^{M_s}$ is sampled i.i.d. from the Gumbel distribution so that $g_{ij}=-\log(-\log u_{ij}))$ where $u_{ij} \sim \mathrm{Uniform}(0, 1)$.
The sigmoid function is defined as usual as
\begin{equation}
    \bm\sigma(\bm v)_i=\frac{\exp(v_i)}{\sum_{j=1}^{M_s}\exp(v_j)}.
\end{equation}

Since the assignment of shapes to parts is \textit{hard}, the softmax formulation is not applicable in our case;
\textit{e.g.}, we cannot have 90\% of one shape and 10\% of another.
We resolve this by using Eq.~\ref{eq:assign_hard} in forward propagation with the gradient of Eq.~\ref{eq:assign_soft} in back propagation, the well-known straight-through Gumbel-softmax estimator~\cite{jang2016categorical}.

\subsection{Object Generation by Assembling Parts}\label{subsec:meth:assumble}
We now cover how the latent poses, primitives, points, and assignment matrix may be generated and assembled into an object (\textit{e.g.} comprising a  point cloud $\bm Y$ and set of parts $\bm \zeta$ in non-canonical coordinates).
$M_s$ primitives (together denoted $\tilde{\bm P}$) are sampled by sampling each of the associated $\eta$ and $\omega$ parameters. 
The assignment matrix $\bm A$ is also sampled along with the poses $\bm\Tau$.  
The canonical object part primitives $\hat{\bm P}$ are then selected with replacement from the canonical candidate shape primitives $\tilde{\bm P}$ via the assignment matrix and transformed by the pose transformations (see \cite{li2021editvae} for transforming a superquadric) to obtain the object primitives in object coordinates, $\bm P$ (see Fig.~\ref{fig:framework} and the notation guide in Appendix A of the supplementary).

Similarly, $M_s$ canonical point clouds are generated by point-decoder neural network modules, and combined via the assignment matrix and part poses to yield
$N = N_p\times M_\Tau$ number of 3D points in the generated point cloud, which we denote by $\bm Y$ (see Fig.~\ref{fig:framework}).

The above steps are shared in both reconstruction and generation, except that the global latent $\bm z_g$ is encoded from the input ground-truth point cloud $\bm X$ in reconstruction and training, whereas it is sampled from the standard Gaussian prior in (unconstrained) generation.

\subsection{Losses}\label{subsec:meth:losses}
We now cover the terms in the reconstruction error of Eq.~\ref{eq:vae}. These are designed to provide an effective inductive bias for unsupervised part segmentation, primitive representation, point reconstruction, and self similarity.

\subsubsection{Points distance $\mathcal{L}_p(\bm X, \bm Y)$} is defined by the mean of parts Chamfer distance as in~\citet{li2021editvae},
\begin{equation}
    \mathcal{L}_p(\bm X, \bm Y) = \sum_{m=1}^{M_\Tau} \mathcal{L}_c(\hat{\bm X}^m, \hat{\bm Y}^m),
\end{equation}
where $\mathcal{L}_c(\cdot)$ denotes Chamfer distance~\cite{li2021editvae}, $\bm X^m$ is subset of the input point cloud $\bm X$ with nearest primitive $\bm P^m$,
and $\hat{\bm X}^m = (\Tau^m)^{-1}(\bm X^m)$ is the same mapped to the canonical coordinate system.
In addition, $\hat{\bm Y}^m \in \bm R^{N_p\times 3}$ is the point cloud output selected by the one-hot vector $\bm A_m$ from $M_s$ points decoders outputs, as in Sec.~\ref{subsec:meth:assign_matrix}.

\subsubsection{Primitives to points distance $\mathcal{L}_r(\bm X, \bm P)$} measures the distance between primitive $\bm P$ and point cloud $\bm X$,
\begin{equation}
    \mathcal{L}_r(\bm X, \bm P) = \mathcal{L}_{\bm P \rightarrow \bm X}(\bm X, \bm P) + \mathcal{L}_{\bm X \rightarrow \bm P}(\bm X, \bm P),
\end{equation}
where $\mathcal{L}_{\bm P \rightarrow \bm X}$ denotes the directed distance from primitives to points, and \textit{vice versa}. We follow~\cite{li2021editvae,paschalidou2019superquadrics} but with $\mathcal{L}_{\bm X \rightarrow \bm P}$ calculated by Eq.~8 rather than Eq.~11 in \cite{paschalidou2019superquadrics}.

\begin{figure*}[pt]
    \begin{center}
        \includegraphics[width=0.97\linewidth]{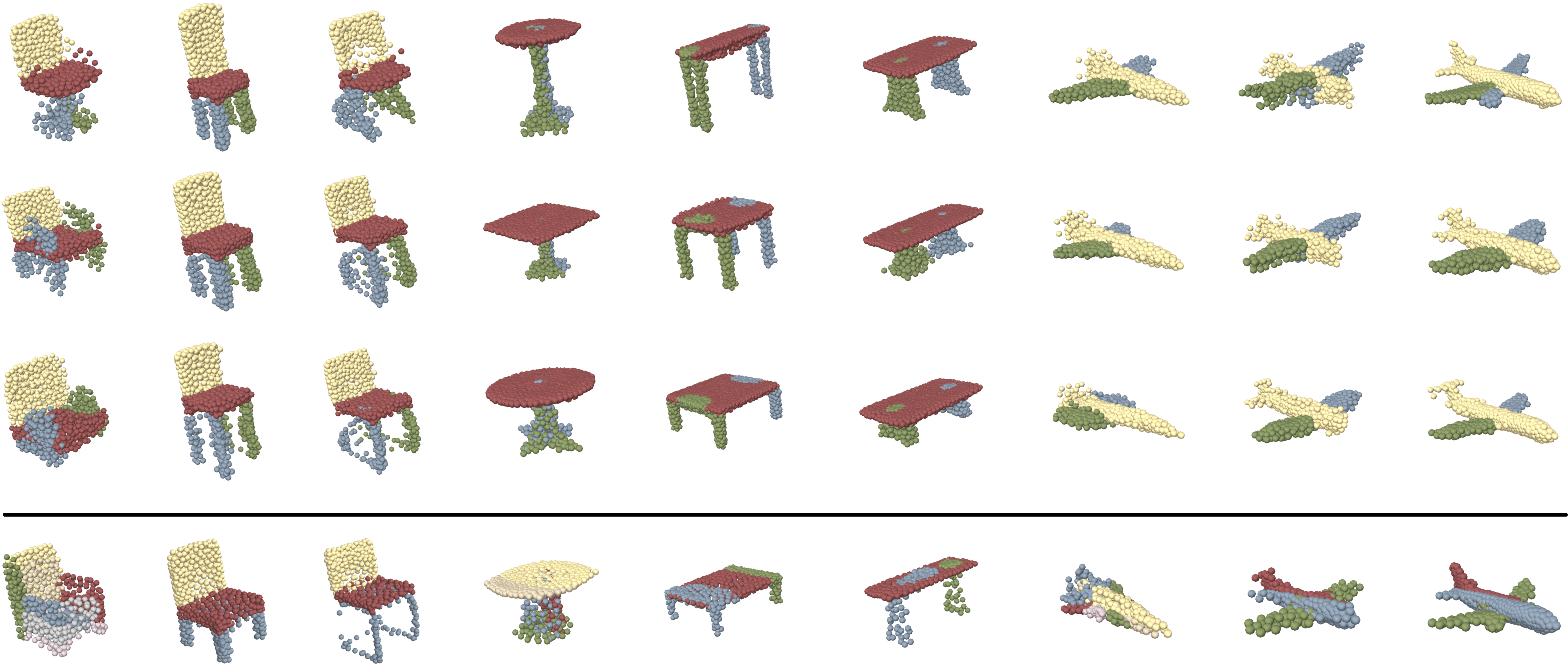}
    \end{center}
       \caption{Part-aware generation for the chair, table, and airplane categories.
       Colored by parts, green and blue parts are mutually similar in all categories, \textit{i.e.} with the same shape but different poses.
       \textit{Top three rows:} examples generated by \spavae. \textit{Bottom row:} examples generated by \editvae\ from the paper~\cite{li2021editvae}.
       }
    \label{fig:generation}
\end{figure*}

\subsubsection{Overlapping loss $\mathcal{L}_o(\bm P)$} discourages primitives from overlapping one another, and is formally defined in terms of the smoothed indicator function $H^m(\cdot)$~\cite{solina1990recovery} as in~\cite{li2021editvae,paschalidou2019superquadrics}.
\begin{align}\label{eq:overlapping}
    &\mathcal{L}_{o}(\bm P) \\
    &\qquad =\frac{1}{M_\Tau} \smash{\sum_{m=1}^{M_\Tau}}
    \frac{1}{\vert \bm S \vert - \vert \bm S^m \vert} \sum_{\substack{\bm y \in \bm S \setminus \bm S^m }} \max\left(s-H^m(\bm y), 0\right) 
    ,
    \nonumber
\end{align}
where $\vert \cdot \vert$ denotes cardinality, $\bm S^m$ denotes a set of points sampled from the surface of $\bm P^m$, and $\bm S = \bigcup_{m=1}^{M_\Tau} \bm S^m$.
In addition, $s$ is a threshold parameter associated with $H^m(\cdot)$ to discourage primitives from overlapping, or even avoid it entirely to achieve the spatial disjointedness.

\subsubsection{Diversity loss $\mathcal{L}_d(\bm \alpha)$} promotes point clouds with highly variable primitive scales.
We use a monotonic squashing transformation of what can be shown to be proportional to the sum of the variances of the components of $\bm \alpha$,  
\begin{align}\label{eq:diversity}
    \mathcal{L}_d(\bm \alpha)
    = \tanh\left(\frac{-c_1\sum_{i=1}^{M_s} \sum_{j=1, j\neq i}^{M_s} \Vert \bm\alpha_i - \bm\alpha_j \Vert_2^2}{M_s(M_s-1)}\right)
    ,
\end{align}
where $\bm \alpha$ is from Eq.~\ref{eq:primitive} and we fixed $c_1=4$ throughout.

\subsubsection{Assignment loss $\mathcal{L}_a(\bm A)$} applies a hinge loss to $\bm A$,
\begin{equation}
    \mathcal{L}_a(\bm A) = \frac{1}{M_s} \sum_{i=1}^{M_s} \max( 1 - \sum_{j=1}^{M_\Tau} \bm A_{ij}, 0),
\end{equation}
thereby encouraging assignment matrix $\bm A$ for which each shape is activated at least once.

\subsection{Implementation and Training Details}\label{subsec:meth:arch_detail}
SPA-VAE is trained via the usual ELBO objective Eq.~\ref{eq:vae}, but in a specific stage-wise fashion that enhances training speed and stability.
In stage 1 we train the primitives for 200 epochs, with the reconstruction error in Eq.~\ref{eq:vae},
\begin{align}\label{eq:recon_loss_prim}
    &\bm E_{Q_\phi(\bm z_g \vert \bm X)}[\log P_\theta(\bm X \vert \bm \zeta)] \\
    &\qquad = \mathcal{L}_r(\bm X, \bm P) + \omega_o \mathcal{L}_o(\bm P) + \omega_d \mathcal{L}_d(\bm\alpha) + \omega_a \mathcal{L}_a(\bm A)
    .
    \nonumber
\end{align}
In stage 2 we train primitives and points together with the reconstruction error
\begin{align}\label{eq:recon_loss_points}
    \bm E_{Q_\phi(\bm z_g \vert \bm X)}[\log P_\theta(\bm X \vert \bm \zeta)]
    = \mathcal{L}_p(\bm X, \bm Y) +
    \omega_a \mathcal{L}_a(\bm A)
    .
    \nonumber
\end{align}

All encoders decoders match~\cite{li2021editvae} except the fully connected  $\bm\Delta_\theta(\bm z_g)$ which generates $\bm A$.

\begin{table*}[h!]
    \centering
    \resizebox{1.0\textwidth}{!}{
    \begin{tabular}{c c l c c c c} 
        \toprule
        Class & Model & JSD $\downarrow$ & MMD-CD $\downarrow$ & MMD-EMD $\downarrow$ & COV-CD $\uparrow$ & COV-EMD $\uparrow$ \\ 
        \midrule
        \multirow{7}{*}{Chair} & \rgan\ (dense)$^\star$ & 0.238 & 0.0029 & 0.136 & 33 & 13 \\
         & \rgan\ (conv)$^\star$ & 0.517 & 0.0030 & 0.223 & 23 & 4 \\
         & Valsesia (no up.)$^\star$ & 0.119 & 0.0033 & 0.104 & 26 & 20 \\
         & Valsesia (up.)$^\star$ & 0.100 & 0.0029 & 0.097 & 30 & 26 \\
         & \treegan\ \cite{shu20193d} & 0.069 & 0.0018 & 0.113 & \textbf{51} & 17 \\
         & \editvae~\cite{li2021editvae} ($M=4$) & \textbf{0.047} & 0.0018 & 0.115 & 45 & \textbf{29} \\
         & \spavae\ ($M_\Tau=4$)& 0.065$^\dagger$ & \textbf{0.0017} & \textbf{0.034} & 39 & 23 \\
         \midrule
         \multirow{7}{*}{Airplane} & \rgan (dense)$^\star$ & 0.182 & 0.0009 & 0.094 & 31 & 9 \\ 
         & \rgan (conv)$^\star$ & 0.350 & 0.0008 & 0.101 & 26 & 7 \\ 
         & Valsesia (no up.)$^\star$ & 0.164 & 0.0010 & 0.102 & 24 & 13 \\
         & Valsesia (up.)$^\star$ & 0.083 & 0.0008 & 0.071 & 31 & 14 \\
         & \treegan\ \cite{shu20193d} & 0.064 & \textbf{0.0004} & 0.070 & \textbf{45} & 9 \\
         & \editvae~\cite{li2021editvae} ($M=3$) & \textbf{0.044} & 0.0005 & 0.067 & 23 & 17 \\
         & \spavae\ ($M_\Tau=3$)& 0.067$^\dagger$ & \textbf{0.0004} & \textbf{0.003} & 39 & \textbf{23} \\
        \midrule
        \multirow{3}{*}{Table} & \treegan\ \cite{shu20193d} & 0.067 & 0.0018 & 0.090 & \textbf{45} & 29 \\
         & \editvae~\cite{li2021editvae} ($M=3$) & \textbf{0.042} & 0.0017 & 0.130 & 39 & \textbf{30} \\
         & \spavae\ ($M_\Tau=3$)& 0.068$^\dagger$ & \textbf{0.0016} & \textbf{0.020} & \textbf{45} & 21 \\
        \bottomrule
       \end{tabular}
    }
    \caption{Generative performance.
    $\uparrow$ means the higher the better, $\downarrow$ means the lower the better.
    The score is highlighted in bold if it is the best one compared with state-of-the-art.
    For network with $\star$ we use the result reported 
    in~\cite{valsesia2018learning,shu20193d,li2021editvae}. JSD scores marked $\dagger$ 
    can be improved significantly by a simple post-processing as per Tab.~\ref{tab:nubalanced_jsd}.
    Both $M_\Tau$ and $M$ represent the number of object parts.
    }
    \label{tab:generative}
\end{table*}
\section{Experiments}
\label{sec:exp}
\subsubsection{Evaluation metrics, baselines, and details.}
\spavae\ is evaluated on the chair, table, and airplane categories of  
ShapeNet~\cite{chang2015shapenet},
with the same data splits and the same evaluation metrics of~\cite{shu20193d,li2021editvae}, \textit{i.e.} JSD, MMD-CD, MMD-EMD, COV-CD, COV-EMD.
We compared our \spavae\ with four existing models: 
r-GAN~\cite{achlioptas2018learning}, Valsesia~\cite{valsesia2018learning}, \treegan~\cite{shu20193d}, and \editvae~\cite{li2021editvae}.
The first of these are baselines which generate point clouds as a whole.
In contrast, \treegan\ and \editvae\ achieve parts-aware point cloud generation. 
Specifically, \treegan\ generates points via a latent tree structure while \editvae\ performs the generation via a disentangled parts-based representation.

All models use input point clouds with 2048 points, and 
infer (global) latent vector representations $\bm z_g \in \bm R^{256}$ with  standard Gaussian prior.
Each generated part consists of 512 points.
\betavae~\cite{burgess2018understanding,higgins2016beta} is adopted in training to obtain a well disentangled latent. We set the dimensionality $\bm z_s \in \bm R^{32}$  for the shape latent and $\bm z_\Tau \in \bm R^{16}$ for the pose latent, respectively. We use
\textsc{Adam}~\cite{kingma2014adam} with  
learning rate 0.0001 for 1000 epochs, a batch size of 30, and no momentum.
Code will be provided on publication of the paper.

\begin{figure}
    \centering
    \begin{subfigure}[b]{0.42\linewidth}
        \centering
       \includegraphics[width=\linewidth]{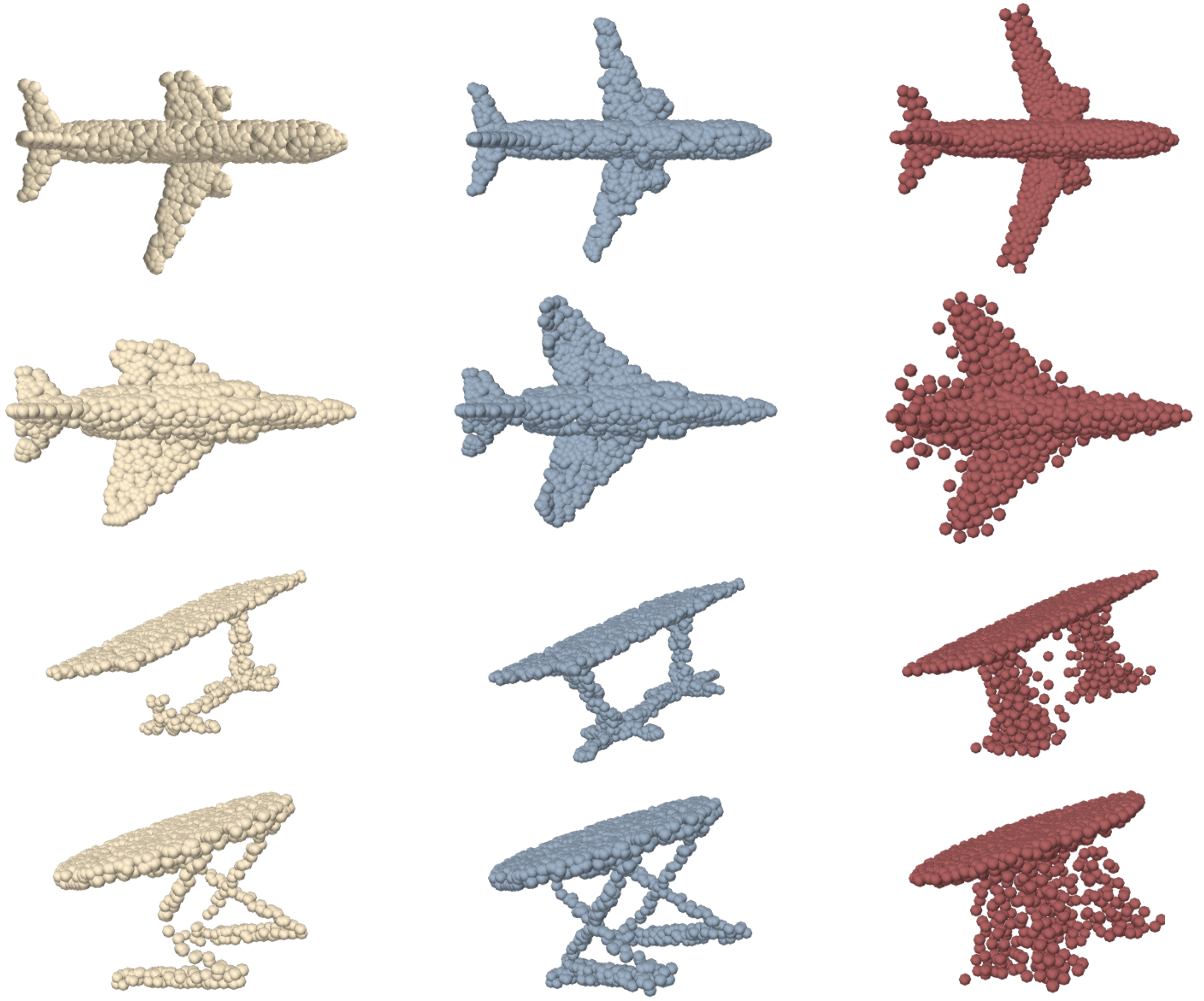}
        \caption{\cut}
    \end{subfigure}
    \begin{subfigure}[b]{0.5\linewidth}
        \centering
       \includegraphics[width=0.76\linewidth]{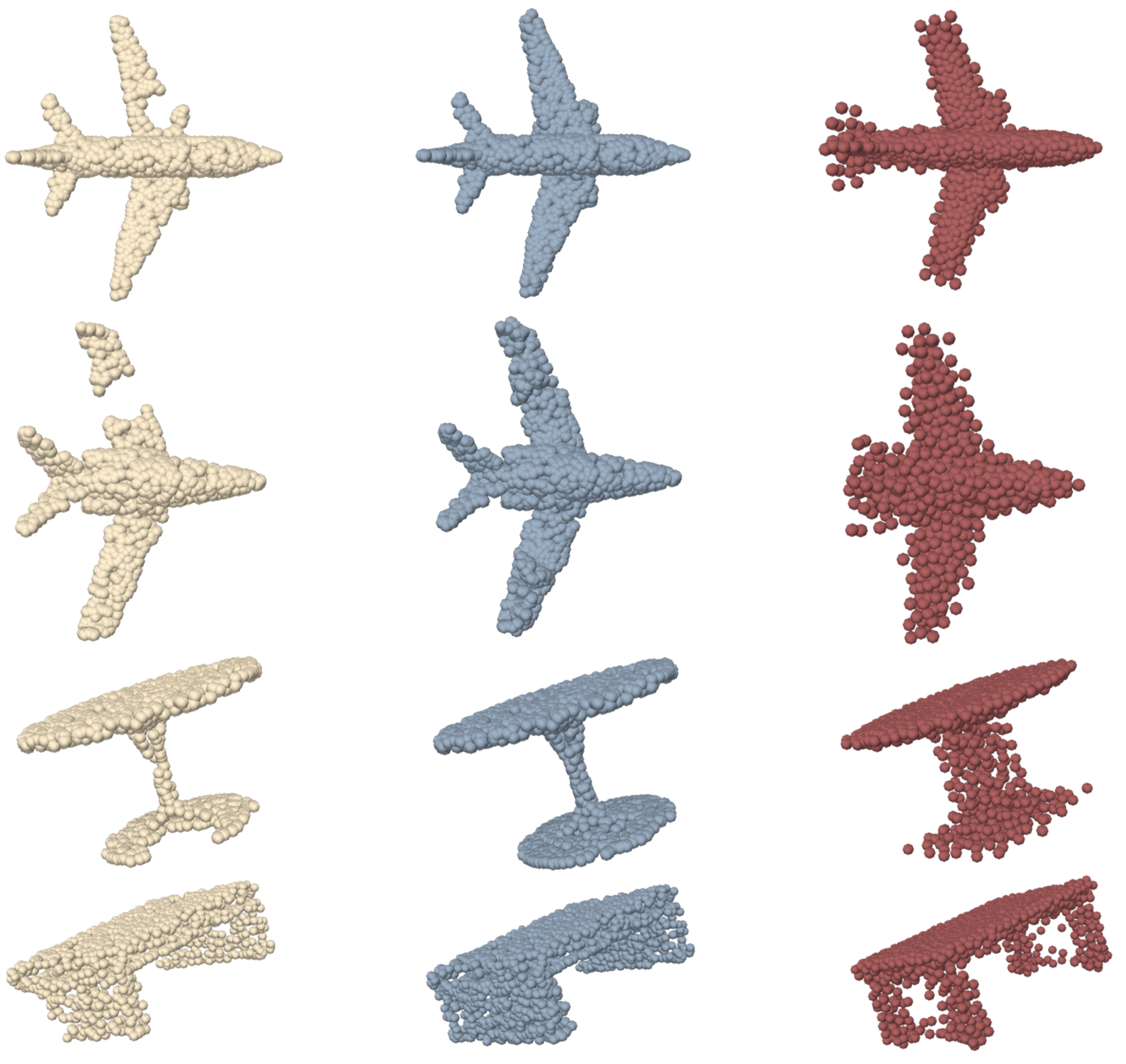}
        \caption{\hole}
    \end{subfigure}
       \caption{
      Completion from the \textit{a)} \cut\ and \textit{b)} \hole\ corruptions.
      \textit{Left:} \uncompleted\ point clouds.
      \textit{Middle:} completion by \completions .
      \textit{Right:} completion by \completionr .
      }
   \label{fig:cutsholes}
\end{figure}
\begin{figure}
    \begin{center}
        \includegraphics[width=\linewidth]{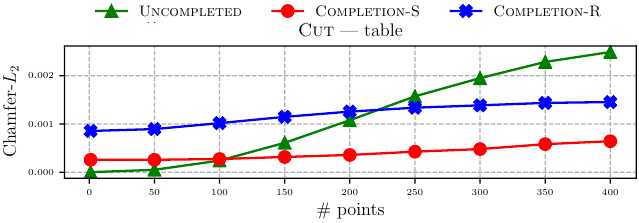}
    \end{center}
       \caption{Completion of \cut\ input point cloud corruptions, for the table category.
       \textit{Horizontal:} the number of points removed from ground truth point cloud.
       \textit{Vertical:} the Chamfer distance between the ground truth point cloud and the \uncompleted\ one (\textit{green}), 
       or the completed one using either the \completionr\ (\textit{blue}), or \completions\ (\textit{red}) methods.
       See Fig.~6 in Appendix D for more categories.
       }
    \label{fig:completion_plot_single}
 \end{figure}
\subsection{Generation}
For point cloud generation we sample the global latent $\bm z_g$, 
and then compute the corresponding poses, primitives and points along with the assignment matrix ---
see Fig.~\ref{fig:framework} for an overview. 
The components thusly generated are assembled to compose the final point cloud output 
in the manner described in Sec.~\ref{subsec:meth:assumble}.
Quantitative and qualitative results are provided in Tab.~\ref{tab:generative} 
and Fig.~\ref{fig:generation}, respectively.
Tab.~\ref{tab:generative} shows that \spavae\ has competitive numerical performance
compared with state-of-the-art approaches. Specifically, 
\spavae\ consistently achieves the best performance under the MMD-EMD and MMD-CD metrics. In line with our intuition, \spavae\ performs especially well on the chair and table categories wherein the \textit{i)} the legs are often poorly sampled and \textit{ii)} the style of a chair or table leg is highly variable across object instances, but highly self-similar within them.
Note that the number of parts (colored in Fig.~\ref{fig:generation}) equals the number of poses $M_\tau$, which 
is manually selected for each category to achieve a semantically meaningful segmentation with potentially similar parts 
in our experiments. To the best of our knowledge, \editvae\ is the closest baseline as 
it also disentangles parts with poses and shapes in an unsupervised manner. But 
it does not explicitly model parts-based self-similarity.
Qualitative visualization and comparisons with \editvae\ are provided in Fig.~\ref{fig:generation}.
\spavae\ achieved better segmentation results as, \textit{e.g.} table legs and airplane
wings are segmented by two similar parts with less part overlap. 
We generally found that, as observed in the figure, \spavae\ generates 3D point clouds with 
more diverse styles, presumably because similar parts from the same object effectively 
augment the training data leading
more detailed and diverse shape prototypes for the parts. For example, the stand and the bottoms of the tables 
are distinguished by more precise details in the fourth column of Fig.~\ref{fig:generation}. Furthermore, \spavae\ tends to avoid mismatched part styles, unlike \textit{e.g.} the arms of the sofa chair generated by \editvae\ at the bottom-left of Fig.~\ref{fig:generation}.

\begin{table}
    \centering
    \resizebox{0.48\textwidth}{!}{
    \begin{tabular}{c|c c c c} 
        \toprule
        Category & \spavae & \spavaeresampled & \editvaeresampled & \editvae \\
        \midrule
        Table & 0.068 & 0.073 & \textbf{0.044} & 0.042 \\
        Chair & 0.065 & 0.047 & \textbf{0.044} & 0.047\\
        Airplane & 0.067 & \textbf{0.032} & 0.044 & 0.044 \\
        \bottomrule
    \end{tabular}
    }
    \captionof{table}{JSD performance for resampled models denoted by the suffix \textsc{-r}.}
    \label{tab:nubalanced_jsd}
\end{table}
\subsection{Completion}
\label{subsec:completion}
The way \spavae\ explicitly learns similarity relationships between parts within an object
provides a new mode of intelligent data-driven 3D shape completion.
We now introduce two variants of \spavae\ to demonstrate the strength of similarity in  
completion tasks. 
Different from recent learning-based completion methods, the demonstration merely uses ground-truth point clouds of complete objects during training,
and incomplete point clouds in testing.

\subsubsection{Completion by Reconstruction.}
As a baseline, completion can be achieved with a pretrained \spavae\ model by simply encoding and decoding the incomplete point cloud. We denote this method by \completionr .

\subsubsection{Completion by Similarity.} 
We give a simple recipe for exploiting the parts-based self-similarity 
inferred by \spavae . 
First we encode the incomplete point cloud 
using the same pretrained \spavae\ model as before. We then
 extract the inferred poses, primitives, and the assignment matrix.
We then assign each point of the incomplete point cloud to the nearest primitive.
Finally, to fill in the missing region of the incomplete point cloud, we copy each input 
point $\bm x$ to each of the corresponding (zero or more) shared parts. This is achieved by 
transforming by $\bm x \mapsto \bm \Tau^{m'} \big((\bm \Tau^{m})^{-1} (\bm x)\big)$ where $m$ and $m'$ are the source and 
target part indices, respectively, yielding a completed point cloud.
We denote this method by \completions . 

We simulate incomplete data with two corruptions. The \cut\ 
corruption removes all the data in a particular half-space of a certain part 
(using \shapenet 's part labels), while the \hole\ corruption removes the 
data inside an $L^2$ ball within a part. We only corrupt
parts with shared (similar) counterparts. 
Further details of these corruption processes may be found in the supplementary materials.
Visualizations of our completions of these corruptions are shown 
in Fig.~\ref{fig:cutsholes}. We also used the Chamfer Distance to measure 
the completion quality quantitatively as summarized in Fig.~\ref{fig:completion_plot_single} and Appendix D Fig.~6.
As we can see, \completions\ outperforms \completionr\ on both the \cut\ 
and \hole\ tasks, in both the airplane and chair categories.
In particular, \completions\ is effective (that is, outperforms 
the \uncompleted\ input point cloud) 
when $\approx 100$ or more points are removed, whereas \completionr\ is effective in this sense only when $\approx 200$ or more point are removed.

See Appendix C for the completion baseline comparsion.

\subsection{Ablation Studies and Measurements}
\subsubsection{Imbalanced Points Distribution}
\label{subsec:imbalanced_pt_dis}
The \spavae\ model can lead to a non-uniform distribution of points per \textit{semantic} part, 
because semantic parts (those corresponding to a fixed real-world definition such as the leg of a chair) tend to be segmented unequally into parts in the sense of the \spavae\ disentanglement. 
This is part of a trade-off between the uniformity of the distribution of points per part, 
and the level of detail captured, since
unbalanced segmentation can help to represent fine details by assigning smaller regions to geometrically complex parts yielding more points per unit of surface area.
The \textit{diversity loss} introduced by \spavae\ helps to model fine details 
(see Fig.~\ref{fig:generation}) but leads to imbalanced points distributions (see 
the supplementary for statistics).
However, the JSD evaluation 
metric\footnote{
The JSD metric typically used on \shapenet\ approximates a point cloud by 
a distribution by quantizing points to a (voxel) grid, counting \& normalizing to a 
probability distribution, and computing the Jensen-Shannon divergence measure between 
two distributions obtained this way.
} 
penalizes non-uniform points distributions because the ground 
truth point clouds tend to be distributed uniformly, and so
may not be suitable for evaluating models aiming at representing fine details more adaptively.

Here we propose a modification to \spavae\ which reduces the impact of  
this imbalance on the JSD metric. 
\spavaeresampled\ (and similarly the baseline \editvaeresampled ) is defined by 
down-sampling the generated parts point cloud by a rate which is set for each 
part separately. The rate is chosen to match the mean number of points assigned 
to the corresponding parts in training.
This yields a final output point cloud which is distributed more uniformly.
As we can see in Tab.~\ref{tab:nubalanced_jsd}, the result is a better
JSD performance, with \spavaeresampled\ comparable to \editvaeresampled\ in terms of JSD.

\subsubsection{Others} See Appendix B for part self-similarity and semantic meaningfulness, as well as the ablation study on novel loss terms.

\section{Conclusion and Limitation}
\label{sec:conclu}
We proposed \spavae , which exploits parts-based self-similarity within the problem of parts-aware generation and completion.
\spavae\ effectively combines the training data from repeated parts of an object, by both segmenting shapes into parts (thereby augmenting the 3D points used for fitting) and explicitly modelling shapes with similar parts (ensuring part consistency).
A limitation of our model is that, due to the naturally existence of self-symmetry within single parts, predicting pose may be challenging and potentially sensitive to noise, especially in the generation setting. We aim to improve this in future works.

\clearpage
\appendix

\onecolumn
\maketitle
\section{Notation Guide}
\label{sec:notation}
\begin{table}[h!]
    \centering
    \resizebox{0.93\textwidth}{!}{
    \begin{tabular}{c|l} 
    \toprule
    Symbol & Explanation \\
    \midrule
    ${\bm X}$ & Input point cloud. \\
    & \\
    $M_s$ & Number of latent candidate shapes. \\
    $M_\Tau$ & Number of parts (and so part poses) in the assembled object. \\
    & \\
    $\bm A$ & \textit{Candidate shape} to \textit{object part} assignment matrix $\in \{0,1\}^{M_s\times M_\Tau}$. \\
    & \\
    $\bm z_g$ & \textit{global} latent embedding vector \\
    $\bm z_\Tau$ & \textit{transformation} (\textit{pose}) latent embedding vector \\
    $\bm z_s$ & \textit{shape} latent embedding vector \\
    & \\
    $\bm \Tau^m$ & $m$-th canonical- to object-coordinate transform, for $m=1,2,\dots M_\Tau$. \\
    & \\
    $\tilde {\bm Y}^m$ & $m$-th candidate shape point cloud in canonical coordinates, for $m=1,2,\dots M_s$. \\
    $\hat {\bm Y}^m$ & $m$-th part point cloud in canonical coordinates, for $m=1,2,\dots M_\Tau$, as selected by $\bm A$. \\
    $ {\bm Y}^m$ & $m$-th part point cloud in object coordinates, for $m=1,2,\dots M_\Tau$, as transformed by $\bm \Tau^m$. \\
    & \\
    $\tilde {\bm P}^m$ & $m$-th candidate shape primitive in canonical coordinates, for $m=1,2,\dots M_s$. \\
    $\hat {\bm P}^m$ & $m$-th part primitive in canonical coordinates, for $m=1,2,\dots M_\Tau$, as selected by $\bm A$. \\
    $ {\bm P}^m$ & $m$-th part primitive in object coordinates, for $m=1,2,\dots M_\Tau$, as transformed by $\bm \Tau^m$. \\
    & \\
    $ \bm \zeta^m$ & $= \{ \bm \Tau^m, \hat{\bm Y}^m, \hat{\bm P}^m \}$, combined $m$-th part representation for $m=1,2,\dots M_\Tau$. \\
    & \\
    ${\bm \zeta}$ & $=\bigcup_{m=1}^{M_\Tau}{\bm \zeta}^m$ combined ${\bm \zeta}^m$. \\
    & \\
    $\tilde{\bm P}$ & $=\bigcup_{m=1}^{M_s}\tilde{\bm P}^m$ combined $\tilde{\bm P}^m$. \\
    $\hat{\bm P}$ & $=\bigcup_{m=1}^{M_\Tau}\hat{\bm P}^m$ combined $\hat{\bm P}^m$. \\
    ${\bm P}$ & $=\bigcup_{m=1}^{M_\Tau}{\bm P}^m$ combined ${\bm P}^m$, the output set of superquadrics. \\
    & \\
    $\tilde{\bm Y}$ & $=\bigcup_{m=1}^{M_s}\tilde{\bm Y}^m$ combined $\tilde{\bm Y}^m$. \\
    $\hat{\bm Y}$ & $=\bigcup_{m=1}^{M_\Tau}\hat{\bm Y}^m$ combined $\hat{\bm Y}^m$. \\
    ${\bm Y}$ & $=\bigcup_{m=1}^{M_\Tau}{\bm Y}^m$ combined ${\bm Y}^m$, the output point cloud. \\
    & \\
    \bottomrule
    \end{tabular}
    }
\end{table}
\twocolumn

\begin{table}[ht]
    \centering
    \resizebox{0.49\textwidth}{!}{
    \begin{tabular}{c|c c c | c c c }
        \toprule
        & \multicolumn{3}{c}{Mean Chamfer Distance} & \multicolumn{3}{|c}{Min Chamfer Distance} \\
        \midrule
        Category & Airplane & Table & Chair & Airplane & Table & Chair \\
        \midrule
        \spavae & \textbf{0.0017} & \textbf{0.0064} & \textbf{0.0023} & \textbf{0.0012} & \textbf{1.3e-5} & \textbf{4.1e-5} \\
        \editvae & 0.0267 & 0.0371 & 0.0251 & 0.0030 & 0.0199 & 0.0143 \\
        \bottomrule
    \end{tabular}
    }
    \captionof{table}{Measurement of parts-based self-similarity.}
    \label{tab:self_sim}
\end{table}
\begin{table}[ht!]
    \centering
    \resizebox{0.45\textwidth}{!}{
    \begin{tabular}{c|c c c}
       \toprule
        & \multicolumn{3}{c}{MCD $\downarrow$} \\
       \cline{2-4}
        & Chair & Airplane & Table \\
        \midrule
       \editvae & 0.0026 & 0.0016 & 0.0121 \\
       \spavae & 0.0058 & 0.0096 & 0.0146 \\
       \textsc{SPA-VAE-m} & \textbf{0.0010} & \textbf{0.0009} & \textbf{0.0011} \\
       \textsc{Neural Parts} & 0.0032 & 0.0018 & 0.0063 \\
       \textsc{BSP-Net} & 0.0102 & 0.0052 & 0.0124 \\
       \spavae\textsc{-m} \textit{-d} & 0.0021 & 0.0015 & 0.0032 \\
       \spavae\textsc{-m} \textit{-a} & 0.0024 & 0.0017 & 0.0028 \\
       \spavae\textsc{-m} \textit{-d} \textit{-a} & 0.0027 & 0.0013 & 0.0035 \\
       \bottomrule
   \end{tabular}
    }
    \caption{Semantic Meaningfulness.
    Here \textit{-d} and \textit{-a} denote training \spavae\ without the diversity loss $\mathcal{L}_d$ and assignment loss $\mathcal{L}_a$, respectively.}
    \label{tab:sem_meaning}
 \end{table}
\section{Ablation Studies and Measurements}
\subsubsection{Parts Self-Similarity and Semantic Meaningfulness}
\textit{Self-similarity} could be measured by the mean or min of Chamfer distances of all pairs of generated parts in canonical position within a single point cloud.
As shown in Tab.~\ref{tab:self_sim}, \spavae\ has smaller mean and min Chamfer distances compared with \editvae, indicating the discovery of self-similar parts.

In Tab.~\ref{tab:sem_meaning}, we measure the \textit{semantic meaningfulness} with MCD introduced in~\cite{li2021editvae}.
We compared the semantic meaningfulness with \editvae~\cite{li2021editvae}, \textsc{Neural Parts}~\cite{paschalidou2021neural}, and \textsc{BSP-Net}~\cite{chen2020bsp}.
We further propose \textsc{SPA-VAE-m} by merging self-similar parts as one, and then calculate MCD.
As \textsc{SPA-VAE-m} strongly outperforms \editvae, we conclude that \spavae\ discovers 3D shapes in a hierarchical manner, which is described below. 
First, it tends to segment 3D shapes by non-shared canonical shapes, whose segmentation is close to semantic ones (proved by good performance of \textsc{SPA-VAE-m}). Second, it further discovers self-similar parts within each discovered part type.

\begin{table}[t]
    \centering
    \resizebox{0.48\textwidth}{!}{
    \begin{tabular}{c c c c c c} 
        \toprule
        Model & JSD $\downarrow$ & MMD-CD $\downarrow$ & MMD-EMD $\downarrow$ & COV-CD $\uparrow$ & COV-EMD $\uparrow$ \\ 
        \midrule
        \spavae\ \textit{-d} & 0.111 & 0.0018 & 0.022 & 44 & 17\\
        \spavae\ \textit{-a} & 0.109 & 0.0018 & 0.118 & 42 & \textbf{23} \\
        \spavae\ \textit{-d} \textit{-a} & 0.114 & 0.0018 & 0.033 & 43 & 17 \\
        \midrule
        \spavae\ & \textbf{0.068} & \textbf{0.0016} & \textbf{0.020} & \textbf{45} & 21 \\
        \bottomrule
    \end{tabular}
    }
    \captionof{table}{Ablation study of our novel losses for the \textit{table} category.
    Here \textit{-d} and \textit{-a} denote training \spavae\ without the diversity loss $\mathcal{L}_d$ and assignment loss $\mathcal{L}_a$, respectively.
    }
    \label{tab:regulariser}
\end{table}
\subsubsection{Ablations Study on Novel Loss Terms}
Recall that we introduced two novel loss terms as part of the \spavae\ model, namely 
the diversity loss $\mathcal{L}_d$ and the assignment 
loss $\mathcal{L}_a$.
We confirmed their importance by removing one, the other, and both from our training 
optimization, yielding the quantitative results shown in Tab.~\ref{tab:regulariser}.

\section{Compare with Completion Baselines}
We compared \completions\ with the State-of-the-art (SOTA) completion method \cite{zhang2021unsupervised} by using their pre-trained model in our incomplete data set directly (``\cite{zhang2021unsupervised} \textsc{Pretrain}''), and training their model with complete data and then applying to incomplete test data (``\cite{zhang2021unsupervised} \textsc{Fair}''), in order to maintain a fair comparison.
We further introduced an 'unfair' setting in which we train the SOTA with complete and incomplete (cut 400 point as in Sec. 4.2) pairs directly (``\cite{zhang2021unsupervised} - 400'').
As the proposed \spavae\ is designed purely using complete point clouds in training, ``\cite{zhang2021unsupervised} - 400'' taking advantage of observing the incomplete data.
All the above models are tested by completing table point clouds cutted by different number of points.
As in Tab.~\ref{tab:comp_baseline}, \completions\ outperforms ``\cite{zhang2021unsupervised} \textsc{Pretrain}'' and ``\cite{zhang2021unsupervised} - 400'', while ``\cite{zhang2021unsupervised} \textsc{Fair}'' failed when only tuning with complete data.
As SOTA completion methods rely heavily on training with both complete and incomplete point clouds pairs, \spavae\ contributes as 1) relevant and diverse incomplete data may be hard to obtain in the real world. 2) transferred models may could not be adopted well. 3) self similarities may allow accurate completion of parts not seen in training yet observed (in a complete self similar part of) the input.
\begin{table*}[h]
    \centering
    \resizebox{0.9\textwidth}{!}{
    \begin{tabular}{c|c c c c c c c c c}
        \toprule
        Number & 1 & 50 & 100 & 150 & 200 & 250 & 300 & 350 & 400  \\
        \midrule
        \uncompleted & $\approx$ \textbf{0} & $\approx$ \textbf{0} & 2.4 & 6.1 & 10.8 & 15.7 & 19.5 & 22.9 & 24.9 \\
        \cite{zhang2021unsupervised} \textsc{Pretrain} & 9.3 & 9.0  & 9.7 & 10.1 & 11.2 & 10.0 & 10.7 & 11.8 & 11.5 \\
        \completions & 2.6 & 2.6 & \textbf{2.7} & \textbf{3.2} & \textbf{3.6} & \textbf{4.3} & \textbf{4.8} & \textbf{5.8} & \textbf{6.4} \\
        \cite{zhang2021unsupervised} - 400 & 7.2 & 6.9 & 7.9 & 8.3 & 9.1 & 8.7 & 9.7 & 10.5 & 9.5\\
        \cite{zhang2021unsupervised} \textsc{Fair} & 821.2 & 823.0 & 794.4 & 789.5 & 839.5 & 841.3 & 810.3 & 839.9 & 860.8 \\
        \bottomrule
    \end{tabular}
    }
    \captionof{table}{Comparison with completion baseline in terms of mean Chamfer distance $\times 10^4$. Category: table, corruption: \cut}
    \label{tab:comp_baseline}
\end{table*}

\section{More Completion Corruption Results}
\begin{figure*}[pt]
    \begin{center}
        \includegraphics[width=\linewidth]{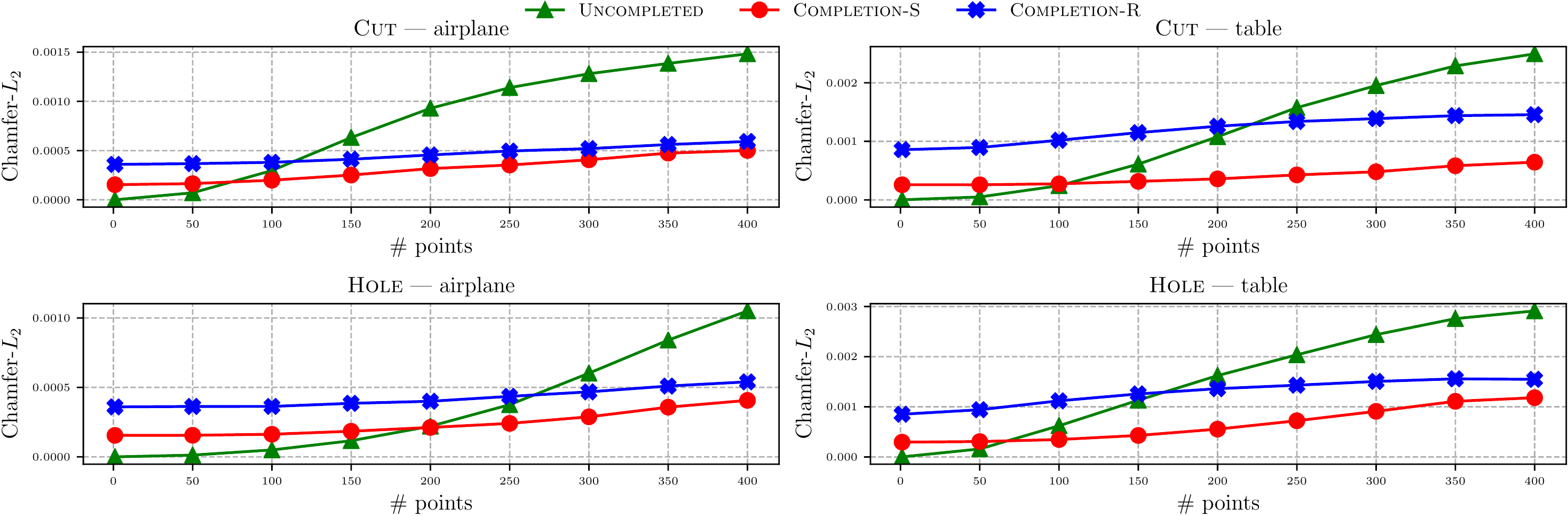}
    \end{center}
       \caption{Completion of \cut\ and \hole\ input point cloud corruptions, for the airplane and table categories.
       \textit{Horizontal:} the number of points removed from ground truth point cloud.
       \textit{Vertical:} the Chamfer distance between the ground truth point cloud and the \uncompleted\ one (\textit{green}), 
       or the completed one using either the \completionr\ (\textit{blue}), or \completions\ (\textit{red}) methods.
       }
    \label{fig:completion_plot}
 \end{figure*}
See Fig~\ref{fig:completion_plot} for more completion corruption results. 

\section{Completion Details}
\label{sec:comp_details}
\subsection{Completion Methods}
\begin{figure*}[pt]
    \begin{center}
        \includegraphics[width=1\linewidth]{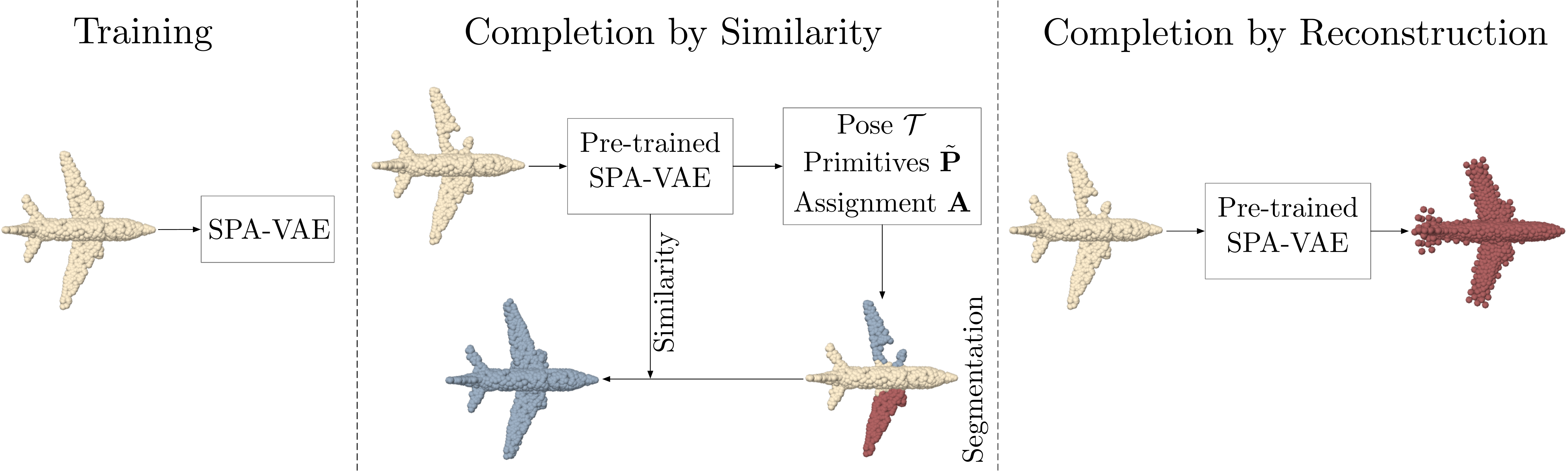}
    \end{center}
       \caption{Visualising completion.
       \textit{Left:} \spavae\ is first pre-trained with complete data.
       \textit{Middle:} Completion by similarity (\completions) is achieved by first inferring an unsupervised segmentation for the incomplete point cloud input using the pre-trained \spavae, and then copying points using the obtained similarity.
       \textit{Right:} Completion by reconstruction (\completionr) is achieved by feeding the incomplete point cloud input into the pre-trained \spavae.
       }
    \label{fig:comp_diagram}
\end{figure*}

Two completion methods (\completionr\ and \completions) are introduced in Sec.~4.2 of the main paper, 
here we further provide the diagram of Fig.~\ref{fig:comp_diagram}. 

\subsection{Corruption Details}
\subsubsection{\cut} corruption is defined by removing all points in a particular half-space of a certain part for which a similar counterpart exists.
The half-space is defined by decreasingly sorting points according the weighted sum of $(x,y,z)$ values and then removing the top $K$ points, with fixed uniformly random weights.

\subsubsection{\hole} corruption is defined by removing a ``ball'' from a certain part whose similar counterpart exists.
The ball is defined by a center point randomly sampled from the part's point cloud, together with $K$ nearest neighborhoods of it.
To distinguish with \cut, we further sample the center point by first sorting all parts points as in \cut, and then selecting the point at the $K$-th rank, to make it more likely that it is interior to the shape.

\section{Visualization of Generated Primitives}
\label{sec:generation+prim}
\begin{figure}[t]
    \begin{center}
        \includegraphics[width=0.8\linewidth]{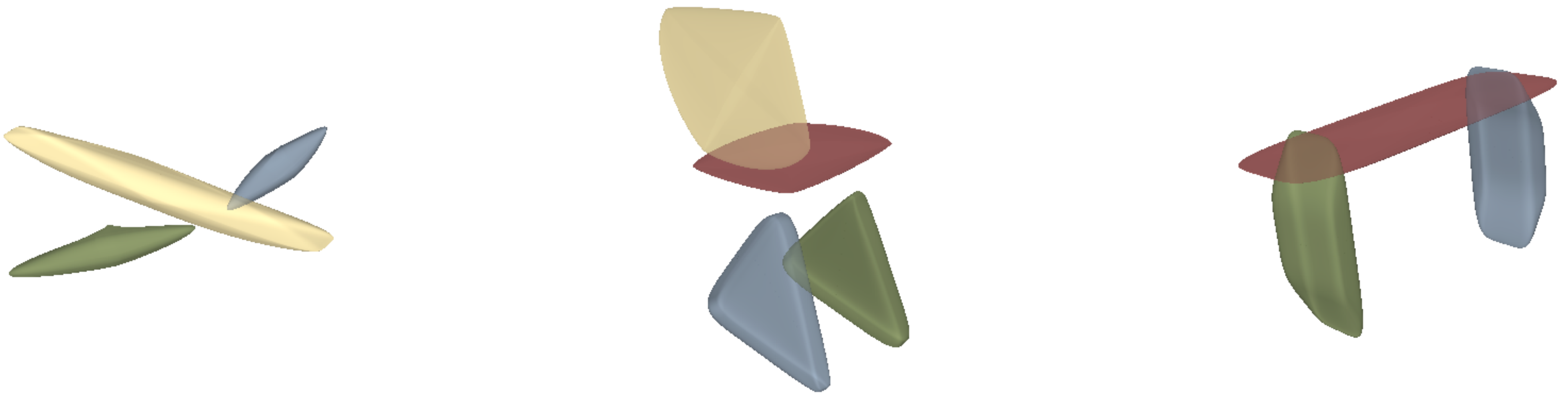}
    \end{center}
       \caption{Generated primitives.
       Each part/primitive is highlighted by different color.
       In all categories, blue and green parts are self-similar parts with shared shape but separate poses.
       }
    \label{fig:primitive}
\end{figure}
The visualization of generated primitives is shown in Fig.~\ref{fig:primitive}, which is assembled from outputs (pose $\Tau$, canonical candidate primitive $\tilde{\bm P}$, and assignment matrix $\bm A$) of a pre-trained \spavae, by sampling from standard Gaussian prior.
See Sec.~3 in the main paper for details.

\section{Another View of Similar Parts Assignment}
\label{sec:spa_view}
We performed similar parts assignment by defining the assignment matrix $\bm A$, which selects shapes (points and primitives) according to Eq.~4 in the main paper.
While the main paper's derivation is clear, here we provide another view of similar parts assignment based on the $n$-mode product~\cite{kolda2009tensor}, to show that the proposed similarity is differentiable, effective and neural-network-friendly.

We take $M_s$ number of \textit{canonical candidate} point clouds $\tilde{\bm Y}^m \in \bm R^{N_p \times 3}$ outputted by point decoders, and concatenate to a tensor $\tilde{{\bm Y}} \in \bm R^{N_p \times 3 \times M_s}$.
Then the selection by assignment matrix $\bm A \in \bm R^{M_s \times M_\Tau}$ is easily expressed as
\begin{align}
    \hat{\bm Y} = \tilde{\bm Y} \times_3 \bm A^T,
\end{align}
here $\times_3$ is 3-mode product. And all point clouds $\hat{\bm Y}^m$ in parts $\bm \zeta^m$ concatenated together to form $\hat{\bm Y} \in \bm R^{N_p \times 3 \times M_\Tau}$.
The column of assignment matrix $\bm A_j$ (row of $\bm A^T$) is a one hot vector, and its dot production with respect to the last dimension of $\hat{\bm Y}$ (number of shapes) simply selects one of $M_s$ shapes, which matches Sec.~2.5 in~\cite{kolda2009tensor}.
Finally, parts point clouds are transformed and composed to form the output point cloud $\bm Y \in \bm R^{N \times 3}$ by way of the parts-based pose transformation function $\bm \Tau^m$ and (inverse) vectorization operation 
as in Sec.~3.3 of the main paper.

Similarly, primitives may be represented by first sampling angles $\eta$ and $\omega$, then transforming to the primitives surface 
with Eq.~3 in the main paper.
Thus, \textit{canonical candidate} primitives $\tilde{\bm P}^m$ may be represented as a set of points on its surface in $\bm R^{N_r \times 3}$, with $N_r$ the number of sampled angles (surface points).
All canonical candidate primitives could be similarly concatenated to form $\tilde{\bm P} \in \bm R^{N_r \times 3 \times M_s}$.
Same as point clouds $\tilde{\bm Y}$, the selection of canonical primitives may be represented as
\begin{align}
    \hat{\bm P} = \tilde{\bm P} \times_3 \bm A^T,
\end{align}
here $\hat{\bm P} \in \bm R^{N_r \times 3 \times M_\Tau}$, which could be separated as primitives $\hat{\bm P}^m$ into parts $\bm \zeta^m$, to finally form the surface-point version of the primitives $\bm P$ by way of poses $\bm \Tau$ and (inverse) vectorization as for the point clouds.

\section{Imbalanced Points Distribution Statistics}
\label{sec:statistics}
\begin{table}[t]
    \centering
    \resizebox{.48\textwidth}{!}{
    \begin{tabular}{c c c c c c c} 
        \toprule
        Model & Category &  \multicolumn{4}{c}{Number of Points Per Part} & \textit{SDev} \\
        \midrule
        \editvae & \multirow{3}{*}{Chair} & 714 & 489 & 445 & 400 & 121\\
        \spavae & & \textbf{756} & \textbf{737} & \textbf{284} & \textbf{271} & \textbf{235} \\
        \balanced & & 512 & 512 & 512 & 512 & 0 \\
        \midrule
        \editvae & \multirow{3}{*}{Table} & 974 & 569 & 505 & - & 208 \\
        \spavae & & \textbf{1234} & \textbf{409} & \textbf{405} & - & \textbf{390}\\
        \balanced & & 683 & 683 & 682 & - & 0\\
        \midrule
        \editvae & \multirow{3}{*}{Airplane} & 813 & 735 & 500 & - & 133 \\
        \spavae & & \textbf{1069} & \textbf{515} & \textbf{464} & - & \textbf{274}\\
        \balanced & & 683 & 683 & 682 & - & 0\\
        \bottomrule
    \end{tabular}
    }
    \caption{Per parts points distribution.
    Here \textit{SDev} denotes the standard deviation of the number of points per part, smaller means closer to uniform points distribution.
    \balanced\ denotes the ideally uniformly distributed points.
    }
    \label{tab:stat}
\end{table}
Here we demonstrate that \spavae\ generates an imbalanced points distribution, which helps to model fine details of point clouds
as described in Sec.~4.3 of the main paper.
The statistics of points distribution are listed in Tab.~\ref{tab:stat}.
Here we assume that ground-truth point clouds used in training and testing have points distributed uniformly.
Ideally, the uniform points distribution in ground-truth point clouds is guaranteed by: 1) each part's point cloud contains the same number of points, and 2) points are distributed uniformly within parts.
The former is reflected by ideally \balanced\ parts generation in Tab.~\ref{tab:stat} and the latter is achieved by points decoders with the help of the Chamfer distance loss, which is not covered further here.

As the parts are designed to output equal numbers of points per points decoder, and imbalanced distribution arises from assigning points in the ground truth input to the nearest part / primitive in reconstruction, for both the \spavae\ and \editvae, in line with Tab.~\ref{tab:stat}.
Because it reflects how much the parts-aware generations violate the above rule (1) --- each generated/segmented parts tend to contain various points number when mapped to the ground-truth point clouds.
Compared with \editvae, \spavae\ outputs parts whose assigned ground-truth points number shows a higher \textit{standard deviation}.
Compared with the ideal represented by \balanced\ whose parts segment the ground truth point cloud equally, high variance means parts generated by \spavae\ tend to contain an imbalanced number of points in the ground truth point cloud, as each part is designed to output an equal number of parts in generation.
Thus,the higher standard deviation exhibited by \spavae\ indicates a more imbalanced points distribution, and so potentially a more detailed representation but worse JSD performance.

\section{Implementation Details}
\label{sec:implement_details}
\subsubsection{Generation Evaluation Metrics}
This paper adopts the same evaluation metrics as in~\cite{achlioptas2018learning,shu20193d,li2021editvae}. To be self-contained, we present their definitions below.
Assume we have a set of $n$ generated point clouds $\bm Y = \{\bm Y_i\} \text{ for } i \in \{1,\dotso,n\}$ and a set of $m$ ground-truth point clouds $\bm X = \{\bm X_j\} \text{ for } j \in \{1,\dotso,m\}$.
\begin{itemize}
    \item \textit{Coverage (COV)} is defined as the fraction of ground-truth point clouds $\bm X$ which is the closest one for any generated point cloud $\bm Y_i$
          \begin{align}
            &\text{COV} (\bm X, \bm Y) = \frac{|\bm X'|}{|\bm X|} \\
            &\bm X' = \{ \bm X_j | \argmax_{\bm X_j \in \bm X} -d(\bm X_j, \bm Y_i) \text{ for } i \in \{1, \dotso, n\} \} \nonumber
          \end{align}
    \item \textit{Minimum Matching Distance (MMD)} is defined as the mean of minimum distance for each ground-truth point cloud $\bm X_j$ with respect to any generated point coud $\bm Y_i$
          \begin{align}
            \text{MMD} (\bm X, \bm Y) = \frac{1}{m} \sum_{j=1}^m \min_{\bm Y_i \in \bm Y} d(\bm X_j, \bm Y_i)
          \end{align}
    \item \textit{Jensen-Shannon Divergence (JSD)} is defined to measure the degree to which the point in axis-aligned generated point clouds $\bm Y$ tends to appear at the similar position as in the ground-truth one $\bm X$.
          As in~\cite{achlioptas2018learning} JSD is defined by a canonical voxel grid in the ambient space.
          With the empirical distribution (counting the number of \textit{all} points within voxels) of the point clouds set defined as $\bm P$, we have
          \begin{align}
            \text{JSD} (P_{\bm X} \Vert P_{\bm Y}) = \frac{1}{2} D(P_{\bm X} \Vert M) + \frac{1}{2} D(P_{\bm Y} \Vert M),
          \end{align}
          where $M = \frac{1}{2}(P_{\bm X} + P_{\bm Y})$ and $D(\cdot \Vert \cdot)$ is the KL-divergence~\cite{kullback1951information}.
\end{itemize}
For the \textit{COV} and \textit{MMD}, $d(\cdot, \cdot)$ is the distance between two point clouds, which could be defined as Chamfer distance and Earth Mover distance~\cite{rubner2000earth}.
The formal definition of the above two distances could be found in the Sec. 2 of~\cite{achlioptas2018learning}.

\subsubsection{Network Architecture}
We adopt the same network architecture as~\cite{li2021editvae} except for the addition of a fully connected layer $\bm \Delta_\theta$ used in the Gumbel softmax estimation 
of Sec.~3.2 of the main paper.
In particular, the encoder is based on \pointnet~\cite{qi2017pointnet}, with \treegan's generator as point decoders \cite{shu20193d}, and fully connected layers as primitive decoders~\cite{paschalidou2019superquadrics}.
We further set $\tau$ in the Gumbel softmax estimator of the main paper~Eq.~6 to 1, which yields a smooth estimator with small gradient.

\subsubsection{Loss Details} See \cite{li2021editvae,paschalidou2019superquadrics} and their supplementary for the details of the points distance $\mathcal{L}_p$, primitives to points distance $\mathcal{L}_r$, and overlapping loss $\mathcal{L}_o$.
We adopt Eq.~8 rather than Eq.~11 in~\cite{paschalidou2019superquadrics} for calculating the directed distance from points to primitives as \spavae\ omits the existence probability otherwise associated with each primitive, in order to simplify the model.
Further, we set $s$ in overlapping loss (Eq.~10 in main paper) as the hyperparameter to control the overlapping or disjointedness among primitives.
Specifically, $s>1$ will promote disjointedness as the value of smoothed inside-outside indicator function~\cite{solina1990recovery} tends to greater than 1, with $s=1$ on the surface, and $s<1$ for overlaps.
We set $c_1=4$ in the diversity loss, to achieve a close loss and gradient value compared with other losses, while keeping away from the convergence.

The weight of losses is assigned to $\omega_o=1e-6$, $\omega_d=1e-6$, $\omega_a=0.1$, and $s=1.3$ in table;
$\omega_o=2e-3$, $\omega_d=3e-3$, $\omega_a=0.1$, and $s=1.5$ in chair;
$\omega_o=1e-3$, $\omega_d=1e-5$, $\omega_a=0.1$, and $s=1$ in airplane.

\subsubsection{Miscellaneous Details}
The model is implemented using \texttt{PyTorch}~\cite{paszke2017automatic} built on \texttt{Ubuntu 16.04}, and trained mainly on one \texttt{GeForce RTX 3090} and one \texttt{GeForce RTX 2080 TI}.
Roughly 5 Gigabytes of GPU memory is allocated, but this depends on the batch size, number of parts, and ShapeNet category.
The selection of the number of parts roughly matches the choice in~\cite{li2021editvae}, except for the chair category whose parts number is 3 in \editvae\ and 4 in \spavae, as the latter further separates the legs into two similar parts.

\bibliography{egbib}

\begin{thebibliography}{88}
\providecommand{\natexlab}[1]{#1}

\bibitem[{Achlioptas et~al.(2018)Achlioptas, Diamanti, Mitliagkas, and
  Guibas}]{achlioptas2018learning}
Achlioptas, P.; Diamanti, O.; Mitliagkas, I.; and Guibas, L. 2018.
\newblock Learning representations and generative models for 3d point clouds.
\newblock In \emph{International conference on machine learning}, 40--49. PMLR.

\bibitem[{Barr(1987)}]{barr1987global}
Barr, A.~H. 1987.
\newblock Global and local deformations of solid primitives.
\newblock In \emph{Readings in Computer Vision}, 661--670. Elsevier.

\bibitem[{Brock, Donahue, and Simonyan(2018)}]{brock2018large}
Brock, A.; Donahue, J.; and Simonyan, K. 2018.
\newblock Large scale GAN training for high fidelity natural image synthesis.
\newblock \emph{arXiv preprint arXiv:1809.11096}.

\bibitem[{Burgess et~al.(2018)Burgess, Higgins, Pal, Matthey, Watters,
  Desjardins, and Lerchner}]{burgess2018understanding}
Burgess, C.~P.; Higgins, I.; Pal, A.; Matthey, L.; Watters, N.; Desjardins, G.;
  and Lerchner, A. 2018.
\newblock Understanding disentangling in $\beta$-VAE.
\newblock \emph{arXiv preprint arXiv:1804.03599}.

\bibitem[{Cai et~al.(2020)Cai, Yang, Averbuch-Elor, Hao, Belongie, Snavely, and
  Hariharan}]{cai2020learning}
Cai, R.; Yang, G.; Averbuch-Elor, H.; Hao, Z.; Belongie, S.; Snavely, N.; and
  Hariharan, B. 2020.
\newblock Learning gradient fields for shape generation.
\newblock In \emph{Computer Vision--ECCV 2020: 16th European Conference,
  Glasgow, UK, August 23--28, 2020, Proceedings, Part III 16}, 364--381.
  Springer.

\bibitem[{Chang et~al.(2015)Chang, Funkhouser, Guibas, Hanrahan, Huang, Li,
  Savarese, Savva, Song, Su et~al.}]{chang2015shapenet}
Chang, A.~X.; Funkhouser, T.; Guibas, L.; Hanrahan, P.; Huang, Q.; Li, Z.;
  Savarese, S.; Savva, M.; Song, S.; Su, H.; et~al. 2015.
\newblock Shapenet: An information-rich 3d model repository.
\newblock \emph{arXiv preprint arXiv:1512.03012}.

\bibitem[{Chen, Chen, and Mitra(2019)}]{chen2019unpaired}
Chen, X.; Chen, B.; and Mitra, N.~J. 2019.
\newblock Unpaired point cloud completion on real scans using adversarial
  training.
\newblock \emph{arXiv preprint arXiv:1904.00069}.

\bibitem[{Chen, Tagliasacchi, and Zhang(2020)}]{chen2020bsp}
Chen, Z.; Tagliasacchi, A.; and Zhang, H. 2020.
\newblock Bsp-net: Generating compact meshes via binary space partitioning.
\newblock In \emph{Proceedings of the IEEE/CVF Conference on Computer Vision
  and Pattern Recognition}, 45--54.

\bibitem[{Chen et~al.(2019)Chen, Yin, Fisher, Chaudhuri, and
  Zhang}]{chen2019bae}
Chen, Z.; Yin, K.; Fisher, M.; Chaudhuri, S.; and Zhang, H. 2019.
\newblock BAE-NET: branched autoencoder for shape co-segmentation.
\newblock In \emph{Proceedings of the IEEE/CVF International Conference on
  Computer Vision}, 8490--8499.

\bibitem[{Deng et~al.(2020)Deng, Genova, Yazdani, Bouaziz, Hinton, and
  Tagliasacchi}]{deng2020cvxnet}
Deng, B.; Genova, K.; Yazdani, S.; Bouaziz, S.; Hinton, G.; and Tagliasacchi,
  A. 2020.
\newblock Cvxnet: Learnable convex decomposition.
\newblock In \emph{Proceedings of the IEEE/CVF Conference on Computer Vision
  and Pattern Recognition}, 31--44.

\bibitem[{Diao et~al.(2021)Diao, Zhang, Ma, and Lu}]{diao2021similarity}
Diao, H.; Zhang, Y.; Ma, L.; and Lu, H. 2021.
\newblock Similarity Reasoning and Filtration for Image-Text Matching.
\newblock Technical report, Technical Report.

\bibitem[{Dubrovina et~al.(2019)Dubrovina, Xia, Achlioptas, Shalah, Groscot,
  and Guibas}]{dubrovina2019composite}
Dubrovina, A.; Xia, F.; Achlioptas, P.; Shalah, M.; Groscot, R.; and Guibas,
  L.~J. 2019.
\newblock Composite shape modeling via latent space factorization.
\newblock In \emph{Proceedings of the IEEE/CVF International Conference on
  Computer Vision}, 8140--8149.

\bibitem[{Gal et~al.(2021)Gal, Bermano, Zhang, and Cohen-Or}]{gal2021mrgan}
Gal, R.; Bermano, A.; Zhang, H.; and Cohen-Or, D. 2021.
\newblock MRGAN: Multi-Rooted 3D Shape Representation Learning With
  Unsupervised Part Disentanglement.
\newblock In \emph{Proceedings of the IEEE/CVF International Conference on
  Computer Vision}, 2039--2048.

\bibitem[{Genova et~al.(2020)Genova, Cole, Sud, Sarna, and
  Funkhouser}]{genova2020local}
Genova, K.; Cole, F.; Sud, A.; Sarna, A.; and Funkhouser, T. 2020.
\newblock Local deep implicit functions for 3d shape.
\newblock In \emph{Proceedings of the IEEE/CVF Conference on Computer Vision
  and Pattern Recognition}, 4857--4866.

\bibitem[{Genova et~al.(2019)Genova, Cole, Vlasic, Sarna, Freeman, and
  Funkhouser}]{genova2019learning}
Genova, K.; Cole, F.; Vlasic, D.; Sarna, A.; Freeman, W.~T.; and Funkhouser, T.
  2019.
\newblock Learning shape templates with structured implicit functions.
\newblock In \emph{Proceedings of the IEEE/CVF International Conference on
  Computer Vision}, 7154--7164.

\bibitem[{Gong et~al.(2021)Gong, Nie, Lin, Han, and Yu}]{gong2021me}
Gong, B.; Nie, Y.; Lin, Y.; Han, X.; and Yu, Y. 2021.
\newblock ME-PCN: Point Completion Conditioned on Mask Emptiness.
\newblock In \emph{Proceedings of the IEEE/CVF International Conference on
  Computer Vision}, 12488--12497.

\bibitem[{Goodfellow et~al.(2014)Goodfellow, Pouget-Abadie, Mirza, Xu,
  Warde-Farley, Ozair, Courville, and Bengio}]{goodfellow2014generative}
Goodfellow, I.; Pouget-Abadie, J.; Mirza, M.; Xu, B.; Warde-Farley, D.; Ozair,
  S.; Courville, A.; and Bengio, Y. 2014.
\newblock Generative adversarial nets.
\newblock \emph{Advances in neural information processing systems}, 27.

\bibitem[{Gu et~al.(2020)Gu, Ma, Manivasagam, Zeng, Wang, Xiong, Su, and
  Urtasun}]{gu2020weakly}
Gu, J.; Ma, W.-C.; Manivasagam, S.; Zeng, W.; Wang, Z.; Xiong, Y.; Su, H.; and
  Urtasun, R. 2020.
\newblock Weakly-supervised 3D Shape Completion in the Wild.
\newblock In \emph{Computer Vision--ECCV 2020: 16th European Conference,
  Glasgow, UK, August 23--28, 2020, Proceedings, Part V 16}, 283--299.
  Springer.

\bibitem[{Higgins et~al.(2016)Higgins, Matthey, Pal, Burgess, Glorot,
  Botvinick, Mohamed, and Lerchner}]{higgins2016beta}
Higgins, I.; Matthey, L.; Pal, A.; Burgess, C.; Glorot, X.; Botvinick, M.;
  Mohamed, S.; and Lerchner, A. 2016.
\newblock beta-vae: Learning basic visual concepts with a constrained
  variational framework.

\bibitem[{Huang and You(2012)}]{huang2012point}
Huang, J.; and You, S. 2012.
\newblock Point cloud matching based on 3D self-similarity.
\newblock In \emph{2012 IEEE Computer Society Conference on Computer Vision and
  Pattern Recognition Workshops}, 41--48. IEEE.

\bibitem[{Huang et~al.(2021)Huang, Zou, Cui, Yang, Wang, Zhao, Zhang, Yuan, Xu,
  and Liu}]{huang2021rfnet}
Huang, T.; Zou, H.; Cui, J.; Yang, X.; Wang, M.; Zhao, X.; Zhang, J.; Yuan, Y.;
  Xu, Y.; and Liu, Y. 2021.
\newblock RFNet: Recurrent Forward Network for Dense Point Cloud Completion.
\newblock In \emph{Proceedings of the IEEE/CVF International Conference on
  Computer Vision}, 12508--12517.

\bibitem[{Hui et~al.(2020)Hui, Xu, Xie, Qian, and Yang}]{hui2020progressive}
Hui, L.; Xu, R.; Xie, J.; Qian, J.; and Yang, J. 2020.
\newblock Progressive point cloud deconvolution generation network.
\newblock In \emph{Computer Vision--ECCV 2020: 16th European Conference,
  Glasgow, UK, August 23--28, 2020, Proceedings, Part XV 16}, 397--413.
  Springer.

\bibitem[{Jang, Gu, and Poole(2016)}]{jang2016categorical}
Jang, E.; Gu, S.; and Poole, B. 2016.
\newblock Categorical reparameterization with gumbel-softmax.
\newblock \emph{arXiv preprint arXiv:1611.01144}.

\bibitem[{Karras, Laine, and Aila(2019)}]{karras2019style}
Karras, T.; Laine, S.; and Aila, T. 2019.
\newblock A style-based generator architecture for generative adversarial
  networks.
\newblock In \emph{Proceedings of the IEEE/CVF Conference on Computer Vision
  and Pattern Recognition}, 4401--4410.

\bibitem[{Kim et~al.(2020)Kim, Lee, Kang, Lee, and Kim}]{kim2020softflow}
Kim, H.; Lee, H.; Kang, W.~H.; Lee, J.~Y.; and Kim, N.~S. 2020.
\newblock Softflow: Probabilistic framework for normalizing flow on manifolds.
\newblock \emph{Advances in Neural Information Processing Systems}, 33.

\bibitem[{Kingma and Ba(2014)}]{kingma2014adam}
Kingma, D.~P.; and Ba, J. 2014.
\newblock Adam: A method for stochastic optimization.
\newblock \emph{arXiv preprint arXiv:1412.6980}.

\bibitem[{Kingma and Welling(2013)}]{kingma2013auto}
Kingma, D.~P.; and Welling, M. 2013.
\newblock Auto-encoding variational bayes.
\newblock \emph{arXiv preprint arXiv:1312.6114}.

\bibitem[{Klokov, Boyer, and Verbeek(2020)}]{klokov2020discrete}
Klokov, R.; Boyer, E.; and Verbeek, J. 2020.
\newblock Discrete point flow networks for efficient point cloud generation.
\newblock In \emph{Computer Vision--ECCV 2020: 16th European Conference,
  Glasgow, UK, August 23--28, 2020, Proceedings, Part XXIII 16}, 694--710.
  Springer.

\bibitem[{Kolda and Bader(2009)}]{kolda2009tensor}
Kolda, T.~G.; and Bader, B.~W. 2009.
\newblock Tensor decompositions and applications.
\newblock \emph{SIAM review}, 51(3): 455--500.

\bibitem[{Kullback and Leibler(1951)}]{kullback1951information}
Kullback, S.; and Leibler, R.~A. 1951.
\newblock On information and sufficiency.
\newblock \emph{The annals of mathematical statistics}, 22(1): 79--86.

\bibitem[{Kwon et~al.(2021)Kwon, Kim, Kwak, and Cho}]{kwon2021learning}
Kwon, H.; Kim, M.; Kwak, S.; and Cho, M. 2021.
\newblock Learning self-similarity in space and time as generalized motion for
  video action recognition.
\newblock In \emph{Proceedings of the IEEE/CVF International Conference on
  Computer Vision}, 13065--13075.

\bibitem[{Li et~al.(2021)Li, Li, Hui, and Fu}]{li2021sp}
Li, R.; Li, X.; Hui, K.-H.; and Fu, C.-W. 2021.
\newblock SP-GAN: sphere-guided 3D shape generation and manipulation.
\newblock \emph{ACM Transactions on Graphics (TOG)}, 40(4): 1--12.

\bibitem[{Li, Liu, and Walder(2022)}]{li2021editvae}
Li, S.; Liu, M.; and Walder, C. 2022.
\newblock EditVAE: Unsupervised Part-Aware Controllable 3D Point Cloud Shape
  Generation.
\newblock In \emph{Proceedings of the AAAI Conference on Artificial
  Intelligence}.

\bibitem[{Luo and Hu(2021)}]{luo2021diffusion}
Luo, S.; and Hu, W. 2021.
\newblock Diffusion probabilistic models for 3d point cloud generation.
\newblock In \emph{Proceedings of the IEEE/CVF Conference on Computer Vision
  and Pattern Recognition}, 2837--2845.

\bibitem[{Maddison, Mnih, and Teh(2016)}]{maddison2016concrete}
Maddison, C.~J.; Mnih, A.; and Teh, Y.~W. 2016.
\newblock The concrete distribution: A continuous relaxation of discrete random
  variables.
\newblock \emph{arXiv preprint arXiv:1611.00712}.

\bibitem[{Mishra et~al.(2021)Mishra, Zhang, Shen, Kumar, Saligrama, and
  Plummer}]{mishra2021effectively}
Mishra, S.; Zhang, Z.; Shen, Y.; Kumar, R.; Saligrama, V.; and Plummer, B.
  2021.
\newblock Effectively Leveraging Attributes for Visual Similarity.
\newblock In \emph{Proceedings of the IEEE/CVF Conference on Computer Vision
  and Pattern Recognition}, 3904--3909.

\bibitem[{Mitra, Guibas, and Pauly(2006)}]{mitra2006partial}
Mitra, N.~J.; Guibas, L.~J.; and Pauly, M. 2006.
\newblock Partial and approximate symmetry detection for 3d geometry.
\newblock \emph{ACM Transactions on Graphics (TOG)}, 25(3): 560--568.

\bibitem[{Mitra et~al.(2013)Mitra, Pauly, Wand, and Ceylan}]{mitra2013symmetry}
Mitra, N.~J.; Pauly, M.; Wand, M.; and Ceylan, D. 2013.
\newblock Symmetry in 3d geometry: Extraction and applications.
\newblock In \emph{Computer Graphics Forum}, volume~32, 1--23. Wiley Online
  Library.

\bibitem[{Mo et~al.(2019)Mo, Guerrero, Yi, Su, Wonka, Mitra, and
  Guibas}]{mo2019structurenet}
Mo, K.; Guerrero, P.; Yi, L.; Su, H.; Wonka, P.; Mitra, N.; and Guibas, L.~J.
  2019.
\newblock Structurenet: Hierarchical graph networks for 3d shape generation.
\newblock \emph{arXiv preprint arXiv:1908.00575}.

\bibitem[{Mo et~al.(2020)Mo, Wang, Yan, and Guibas}]{mo2020pt2pc}
Mo, K.; Wang, H.; Yan, X.; and Guibas, L. 2020.
\newblock PT2PC: Learning to generate 3d point cloud shapes from part tree
  conditions.
\newblock In \emph{European Conference on Computer Vision}, 683--701. Springer.

\bibitem[{Nash and Williams(2017)}]{nash2017shape}
Nash, C.; and Williams, C.~K. 2017.
\newblock The shape variational autoencoder: A deep generative model of
  part-segmented 3D objects.
\newblock In \emph{Computer Graphics Forum}, volume~36, 1--12. Wiley Online
  Library.

\bibitem[{Nie et~al.(2020)Nie, Lin, Han, Guo, Chang, Cui, and
  Zhang}]{nie2020skeleton}
Nie, Y.; Lin, Y.; Han, X.; Guo, S.; Chang, J.; Cui, S.; and Zhang, J.~J. 2020.
\newblock Skeleton-bridged point completion: From global inference to local
  adjustment.
\newblock \emph{arXiv preprint arXiv:2010.07428}.

\bibitem[{Paschalidou, Gool, and Geiger(2020)}]{paschalidou2020learning}
Paschalidou, D.; Gool, L.~V.; and Geiger, A. 2020.
\newblock Learning unsupervised hierarchical part decomposition of 3d objects
  from a single rgb image.
\newblock In \emph{Proceedings of the IEEE/CVF Conference on Computer Vision
  and Pattern Recognition}, 1060--1070.

\bibitem[{Paschalidou et~al.(2021)Paschalidou, Katharopoulos, Geiger, and
  Fidler}]{paschalidou2021neural}
Paschalidou, D.; Katharopoulos, A.; Geiger, A.; and Fidler, S. 2021.
\newblock Neural Parts: Learning expressive 3D shape abstractions with
  invertible neural networks.
\newblock In \emph{Proceedings of the IEEE/CVF Conference on Computer Vision
  and Pattern Recognition}, 3204--3215.

\bibitem[{Paschalidou, Ulusoy, and Geiger(2019)}]{paschalidou2019superquadrics}
Paschalidou, D.; Ulusoy, A.~O.; and Geiger, A. 2019.
\newblock Superquadrics revisited: Learning 3d shape parsing beyond cuboids.
\newblock In \emph{Proceedings of the IEEE/CVF Conference on Computer Vision
  and Pattern Recognition}, 10344--10353.

\bibitem[{Paszke et~al.(2017)Paszke, Gross, Chintala, Chanan, Yang, DeVito,
  Lin, Desmaison, Antiga, and Lerer}]{paszke2017automatic}
Paszke, A.; Gross, S.; Chintala, S.; Chanan, G.; Yang, E.; DeVito, Z.; Lin, Z.;
  Desmaison, A.; Antiga, L.; and Lerer, A. 2017.
\newblock Automatic differentiation in pytorch.

\bibitem[{Pauly et~al.(2008)Pauly, Mitra, Wallner, Pottmann, and
  Guibas}]{pauly2008discovering}
Pauly, M.; Mitra, N.~J.; Wallner, J.; Pottmann, H.; and Guibas, L.~J. 2008.
\newblock Discovering structural regularity in 3D geometry.
\newblock In \emph{ACM SIGGRAPH 2008 papers}, 1--11.

\bibitem[{Podolak et~al.(2006)Podolak, Shilane, Golovinskiy, Rusinkiewicz, and
  Funkhouser}]{podolak2006planar}
Podolak, J.; Shilane, P.; Golovinskiy, A.; Rusinkiewicz, S.; and Funkhouser, T.
  2006.
\newblock A planar-reflective symmetry transform for 3D shapes.
\newblock In \emph{ACM SIGGRAPH 2006 Papers}, 549--559.

\bibitem[{Postels et~al.(2021)Postels, Liu, Spezialetti, Van~Gool, and
  Tombari}]{postels2021go}
Postels, J.; Liu, M.; Spezialetti, R.; Van~Gool, L.; and Tombari, F. 2021.
\newblock Go with the Flows: Mixtures of Normalizing Flows for Point Cloud
  Generation and Reconstruction.
\newblock \emph{arXiv preprint arXiv:2106.03135}.

\bibitem[{Qi et~al.(2017)Qi, Su, Mo, and Guibas}]{qi2017pointnet}
Qi, C.~R.; Su, H.; Mo, K.; and Guibas, L.~J. 2017.
\newblock Pointnet: Deep learning on point sets for 3d classification and
  segmentation.
\newblock In \emph{Proceedings of the IEEE conference on computer vision and
  pattern recognition}, 652--660.

\bibitem[{Rezende, Mohamed, and Wierstra(2014)}]{pmlr-v32-rezende14}
Rezende, D.~J.; Mohamed, S.; and Wierstra, D. 2014.
\newblock Stochastic Backpropagation and Approximate Inference in Deep
  Generative Models.
\newblock In Xing, E.~P.; and Jebara, T., eds., \emph{Proceedings of the 31st
  International Conference on Machine Learning}, volume~32 of \emph{Proceedings
  of Machine Learning Research}, 1278--1286. Bejing, China: PMLR.

\bibitem[{Rubner, Tomasi, and Guibas(2000)}]{rubner2000earth}
Rubner, Y.; Tomasi, C.; and Guibas, L.~J. 2000.
\newblock The earth mover's distance as a metric for image retrieval.
\newblock \emph{International journal of computer vision}, 40(2): 99--121.

\bibitem[{Schor et~al.(2019)Schor, Katzir, Zhang, and
  Cohen-Or}]{schor2019componet}
Schor, N.; Katzir, O.; Zhang, H.; and Cohen-Or, D. 2019.
\newblock Componet: Learning to generate the unseen by part synthesis and
  composition.
\newblock In \emph{Proceedings of the IEEE/CVF International Conference on
  Computer Vision}, 8759--8768.

\bibitem[{Seo, Shim, and Cho(2021)}]{seo2021learning}
Seo, A.; Shim, W.; and Cho, M. 2021.
\newblock Learning to Discover Reflection Symmetry via Polar Matching
  Convolution.
\newblock In \emph{Proceedings of the IEEE/CVF International Conference on
  Computer Vision}, 1285--1294.

\bibitem[{Shu, Park, and Kwon(2019)}]{shu20193d}
Shu, D.~W.; Park, S.~W.; and Kwon, J. 2019.
\newblock 3d point cloud generative adversarial network based on tree
  structured graph convolutions.
\newblock In \emph{Proceedings of the IEEE/CVF International Conference on
  Computer Vision}, 3859--3868.

\bibitem[{Sipiran, Gregor, and Schreck(2014)}]{sipiran2014approximate}
Sipiran, I.; Gregor, R.; and Schreck, T. 2014.
\newblock Approximate symmetry detection in partial 3d meshes.
\newblock In \emph{Computer Graphics Forum}, volume~33, 131--140. Wiley Online
  Library.

\bibitem[{Solina and Bajcsy(1990)}]{solina1990recovery}
Solina, F.; and Bajcsy, R. 1990.
\newblock Recovery of parametric models from range images: The case for
  superquadrics with global deformations.
\newblock \emph{IEEE transactions on pattern analysis and machine
  intelligence}, 12(2): 131--147.

\bibitem[{Stutz and Geiger(2018)}]{stutz2018learning}
Stutz, D.; and Geiger, A. 2018.
\newblock Learning 3d shape completion from laser scan data with weak
  supervision.
\newblock In \emph{Proceedings of the IEEE Conference on Computer Vision and
  Pattern Recognition}, 1955--1964.

\bibitem[{Sun et~al.(2019)Sun, Zou, Tong, and Liu}]{sun2019learning}
Sun, C.-Y.; Zou, Q.-F.; Tong, X.; and Liu, Y. 2019.
\newblock Learning adaptive hierarchical cuboid abstractions of 3d shape
  collections.
\newblock \emph{ACM Transactions on Graphics (TOG)}, 38(6): 1--13.

\bibitem[{Sung et~al.(2015)Sung, Kim, Angst, and Guibas}]{sung2015data}
Sung, M.; Kim, V.~G.; Angst, R.; and Guibas, L. 2015.
\newblock Data-driven structural priors for shape completion.
\newblock \emph{ACM Transactions on Graphics (TOG)}, 34(6): 1--11.

\bibitem[{Tchapmi et~al.(2019)Tchapmi, Kosaraju, Rezatofighi, Reid, and
  Savarese}]{tchapmi2019topnet}
Tchapmi, L.~P.; Kosaraju, V.; Rezatofighi, H.; Reid, I.; and Savarese, S. 2019.
\newblock Topnet: Structural point cloud decoder.
\newblock In \emph{Proceedings of the IEEE/CVF Conference on Computer Vision
  and Pattern Recognition}, 383--392.

\bibitem[{Thrun and Wegbreit(2005)}]{thrun2005shape}
Thrun, S.; and Wegbreit, B. 2005.
\newblock Shape from symmetry.
\newblock In \emph{Tenth IEEE International Conference on Computer Vision
  (ICCV'05) Volume 1}, volume~2, 1824--1831. IEEE.

\bibitem[{Tulsiani et~al.(2017)Tulsiani, Su, Guibas, Efros, and
  Malik}]{tulsiani2017learning}
Tulsiani, S.; Su, H.; Guibas, L.~J.; Efros, A.~A.; and Malik, J. 2017.
\newblock Learning shape abstractions by assembling volumetric primitives.
\newblock In \emph{Proceedings of the IEEE Conference on Computer Vision and
  Pattern Recognition}, 2635--2643.

\bibitem[{Valsesia, Fracastoro, and Magli(2018)}]{valsesia2018learning}
Valsesia, D.; Fracastoro, G.; and Magli, E. 2018.
\newblock Learning localized generative models for 3d point clouds via graph
  convolution.
\newblock In \emph{International conference on learning representations}.

\bibitem[{Walder and Kim(2018{\natexlab{a}})}]{pmlr-v80-walder18a}
Walder, C.; and Kim, D. 2018{\natexlab{a}}.
\newblock Neural Dynamic Programming for Musical Self Similarity.
\newblock In Dy, J.; and Krause, A., eds., \emph{Proceedings of the 35th
  International Conference on Machine Learning}, volume~80 of \emph{Proceedings
  of Machine Learning Research}, 5105--5113. PMLR.

\bibitem[{Walder and Kim(2018{\natexlab{b}})}]{walder2018neural}
Walder, C.; and Kim, D. 2018{\natexlab{b}}.
\newblock Neural dynamic programming for musical self similarity.
\newblock In \emph{International Conference on Machine Learning}, 5105--5113.
  PMLR.

\bibitem[{Wang, Ang, and Lee(2021)}]{wang2021voxel}
Wang, X.; Ang, M.~H.; and Lee, G.~H. 2021.
\newblock Voxel-based Network for Shape Completion by Leveraging Edge
  Generation.
\newblock In \emph{Proceedings of the IEEE/CVF International Conference on
  Computer Vision}, 13189--13198.

\bibitem[{Wang, Ang~Jr, and Lee(2020{\natexlab{a}})}]{wang2020cascaded}
Wang, X.; Ang~Jr, M.~H.; and Lee, G.~H. 2020{\natexlab{a}}.
\newblock Cascaded refinement network for point cloud completion.
\newblock In \emph{Proceedings of the IEEE/CVF Conference on Computer Vision
  and Pattern Recognition}, 790--799.

\bibitem[{Wang, Ang~Jr, and Lee(2020{\natexlab{b}})}]{wang2020self}
Wang, X.; Ang~Jr, M.~H.; and Lee, G.~H. 2020{\natexlab{b}}.
\newblock A Self-supervised Cascaded Refinement Network for Point Cloud
  Completion.
\newblock \emph{arXiv preprint arXiv:2010.08719}.

\bibitem[{Wen, Yu, and Tao(2021)}]{wen2021learning}
Wen, C.; Yu, B.; and Tao, D. 2021.
\newblock Learning Progressive Point Embeddings for 3D Point Cloud Generation.
\newblock In \emph{Proceedings of the IEEE/CVF Conference on Computer Vision
  and Pattern Recognition}, 10266--10275.

\bibitem[{Wen et~al.(2021{\natexlab{a}})Wen, Han, Cao, Wan, Zheng, and
  Liu}]{wen2021cycle4completion}
Wen, X.; Han, Z.; Cao, Y.-P.; Wan, P.; Zheng, W.; and Liu, Y.-S.
  2021{\natexlab{a}}.
\newblock Cycle4completion: Unpaired point cloud completion using cycle
  transformation with missing region coding.
\newblock In \emph{Proceedings of the IEEE/CVF Conference on Computer Vision
  and Pattern Recognition}, 13080--13089.

\bibitem[{Wen et~al.(2021{\natexlab{b}})Wen, Xiang, Han, Cao, Wan, Zheng, and
  Liu}]{wen2021pmp}
Wen, X.; Xiang, P.; Han, Z.; Cao, Y.-P.; Wan, P.; Zheng, W.; and Liu, Y.-S.
  2021{\natexlab{b}}.
\newblock Pmp-net: Point cloud completion by learning multi-step point moving
  paths.
\newblock In \emph{Proceedings of the IEEE/CVF Conference on Computer Vision
  and Pattern Recognition}, 7443--7452.

\bibitem[{Wu et~al.(2020)Wu, Chen, Zhuang, and Chen}]{wu2020multimodal}
Wu, R.; Chen, X.; Zhuang, Y.; and Chen, B. 2020.
\newblock Multimodal shape completion via conditional generative adversarial
  networks.
\newblock In \emph{Computer Vision--ECCV 2020: 16th European Conference,
  Glasgow, UK, August 23--28, 2020, Proceedings, Part IV 16}, 281--296.
  Springer.

\bibitem[{Wu, Rupprecht, and Vedaldi(2020)}]{wu2020unsupervised}
Wu, S.; Rupprecht, C.; and Vedaldi, A. 2020.
\newblock Unsupervised learning of probably symmetric deformable 3d objects
  from images in the wild.
\newblock In \emph{Proceedings of the IEEE/CVF Conference on Computer Vision
  and Pattern Recognition}, 1--10.

\bibitem[{Xiang et~al.(2021)Xiang, Wen, Liu, Cao, Wan, Zheng, and
  Han}]{xiang2021snowflakenet}
Xiang, P.; Wen, X.; Liu, Y.-S.; Cao, Y.-P.; Wan, P.; Zheng, W.; and Han, Z.
  2021.
\newblock Snowflakenet: Point cloud completion by snowflake point deconvolution
  with skip-transformer.
\newblock In \emph{Proceedings of the IEEE/CVF International Conference on
  Computer Vision}, 5499--5509.

\bibitem[{Xiao et~al.(2021)Xiao, Reed, Wang, Keutzer, and
  Darrell}]{xiao2021region}
Xiao, T.; Reed, C.~J.; Wang, X.; Keutzer, K.; and Darrell, T. 2021.
\newblock Region similarity representation learning.
\newblock \emph{arXiv preprint arXiv:2103.12902}.

\bibitem[{Xie et~al.(2021)Xie, Wang, Zhang, Yang, Chen, and Wen}]{xie2021style}
Xie, C.; Wang, C.; Zhang, B.; Yang, H.; Chen, D.; and Wen, F. 2021.
\newblock Style-based Point Generator with Adversarial Rendering for Point
  Cloud Completion.
\newblock In \emph{Proceedings of the IEEE/CVF Conference on Computer Vision
  and Pattern Recognition}, 4619--4628.

\bibitem[{Xie et~al.(2020)Xie, Yao, Zhou, Mao, Zhang, and Sun}]{xie2020grnet}
Xie, H.; Yao, H.; Zhou, S.; Mao, J.; Zhang, S.; and Sun, W. 2020.
\newblock Grnet: Gridding residual network for dense point cloud completion.
\newblock In \emph{European Conference on Computer Vision}, 365--381. Springer.

\bibitem[{Xu et~al.(2020)Xu, Fan, Yuan, and Singh}]{xu2020ladybird}
Xu, Y.; Fan, T.; Yuan, Y.; and Singh, G. 2020.
\newblock Ladybird: Quasi-monte carlo sampling for deep implicit field based 3d
  reconstruction with symmetry.
\newblock In \emph{European Conference on Computer Vision}, 248--263. Springer.

\bibitem[{Yang et~al.(2019{\natexlab{a}})Yang, Huang, Hao, Liu, Belongie, and
  Hariharan}]{yang2019pointflow}
Yang, G.; Huang, X.; Hao, Z.; Liu, M.-Y.; Belongie, S.; and Hariharan, B.
  2019{\natexlab{a}}.
\newblock Pointflow: 3d point cloud generation with continuous normalizing
  flows.
\newblock In \emph{Proceedings of the IEEE/CVF International Conference on
  Computer Vision}, 4541--4550.

\bibitem[{Yang et~al.(2019{\natexlab{b}})Yang, Zhang, Ni, Li, Liu, Zhou, and
  Tian}]{yang2019modeling}
Yang, J.; Zhang, Q.; Ni, B.; Li, L.; Liu, J.; Zhou, M.; and Tian, Q.
  2019{\natexlab{b}}.
\newblock Modeling point clouds with self-attention and gumbel subset sampling.
\newblock In \emph{Proceedings of the IEEE/CVF Conference on Computer Vision
  and Pattern Recognition}, 3323--3332.

\bibitem[{Yang and Chen(2021)}]{yang2021unsupervised}
Yang, K.; and Chen, X. 2021.
\newblock Unsupervised learning for cuboid shape abstraction via joint
  segmentation from point clouds.
\newblock \emph{ACM Transactions on Graphics (TOG)}, 40(4): 1--11.

\bibitem[{Yang et~al.(2021)Yang, Wu, Zhang, and Jin}]{yang2021cpcgan}
Yang, X.; Wu, Y.; Zhang, K.; and Jin, C. 2021.
\newblock CPCGAN: A Controllable 3D Point Cloud Generative Adversarial Network
  with Semantic Label Generating.
\newblock In \emph{Proceedings of the AAAI Conference on Artificial
  Intelligence}, volume~35, 3154--3162.

\bibitem[{Yoon et~al.(2020)Yoon, Kang, Jeon, Lee, Han, Park, and
  Kim}]{yoon2020image}
Yoon, S.; Kang, W.~Y.; Jeon, S.; Lee, S.; Han, C.; Park, J.; and Kim, E.-S.
  2020.
\newblock Image-to-Image Retrieval by Learning Similarity between Scene Graphs.
\newblock \emph{arXiv preprint arXiv:2012.14700}.

\bibitem[{Yuan et~al.(2018)Yuan, Khot, Held, Mertz, and Hebert}]{yuan2018pcn}
Yuan, W.; Khot, T.; Held, D.; Mertz, C.; and Hebert, M. 2018.
\newblock Pcn: Point completion network.
\newblock In \emph{2018 International Conference on 3D Vision (3DV)}, 728--737.
  IEEE.

\bibitem[{Zhang et~al.(2021)Zhang, Chen, Cai, Pan, Zhao, Yi, Yeo, Dai, and
  Loy}]{zhang2021unsupervised}
Zhang, J.; Chen, X.; Cai, Z.; Pan, L.; Zhao, H.; Yi, S.; Yeo, C.~K.; Dai, B.;
  and Loy, C.~C. 2021.
\newblock Unsupervised 3D Shape Completion through GAN Inversion.
\newblock In \emph{Proceedings of the IEEE/CVF Conference on Computer Vision
  and Pattern Recognition}, 1768--1777.

\bibitem[{Zhou, Du, and Wu(2021)}]{zhou20213d}
Zhou, L.; Du, Y.; and Wu, J. 2021.
\newblock 3d shape generation and completion through point-voxel diffusion.
\newblock \emph{arXiv preprint arXiv:2104.03670}.

\bibitem[{Zhou, Liu, and Ma(2021)}]{zhou2021nerd}
Zhou, Y.; Liu, S.; and Ma, Y. 2021.
\newblock NeRD: Neural 3D Reflection Symmetry Detector.
\newblock In \emph{Proceedings of the IEEE/CVF Conference on Computer Vision
  and Pattern Recognition}, 15940--15949.

\end{thebibliography}

\end{document}